\documentclass{article}

\usepackage[final]{neurips_2025}

\usepackage[utf8]{inputenc} %
\usepackage[T1]{fontenc}    %
\usepackage{hyperref}       %
\usepackage{url}            %
\usepackage{booktabs}       %
\usepackage{amsfonts}       %
\usepackage{nicefrac}       %
\usepackage{microtype}      %
\usepackage{xcolor}         %

\usepackage[capitalise]{cleveref}
\usepackage{graphicx}
\usepackage{wrapfig}
\usepackage{placeins}
\usepackage{xcolor}
\usepackage{todonotes}

\title{From Dormant to Deleted: Tamper-Resistant Unlearning Through Weight-Space Regularization}

\author{ Shoaib Ahmed Siddiqui \thanks{Correspondence to
\url{msas3@cam.ac.uk}.} \\
University of Cambridge
\And
Adrian Weller \\
University of Cambridge \\ The Alan Turing Institute
\And
David Krueger \\
Mila
\AND
Gintare Karolina Dziugaite \\
Google DeepMind \\ Mila
\And
Michael C. Mozer \\
Google DeepMind
\And
Eleni Triantafillou \\
Google DeepMind
}

\begin{document}
\newcommand{\mcm}[1]{\todo[backgroundcolor=white, textcolor=blue, bordercolor=white, author=]{\tiny MCM: {#1}}}
\newcommand{\sas}[1]{\todo[backgroundcolor=white, textcolor=red, author=SAS, caption=SAS]{\tiny {#1}}}

\maketitle

\begin{abstract}
Recent unlearning methods for LLMs are vulnerable to relearning attacks: knowledge believed-to-be-unlearned re-emerges by fine-tuning on a small set of (even seemingly-unrelated) examples. We study this phenomenon in a controlled setting for example-level unlearning in vision classifiers. We make the surprising discovery that forget-set accuracy can recover from around 50\% post-unlearning to nearly 100\% with fine-tuning on just the \textit{retain} set---i.e., zero examples of the forget set. We observe this effect across a wide variety of unlearning methods, whereas for a model retrained from scratch excluding the forget set (gold standard), the accuracy remains at 50\%. We observe that resistance to relearning attacks can be predicted by weight-space properties, specifically, $L_2$-distance and linear mode connectivity between the original and the unlearned model. Leveraging this insight, we propose a new class of methods that achieve state-of-the-art resistance to relearning attacks\footnote{Code to reproduce our experiments: \url{https://github.com/shoaibahmed/vision_relearning}}.
\end{abstract}

\section{Introduction}

\begin{wrapfigure}{r}{0.5\textwidth}
    \centering
    \vspace{-5mm}
    \includegraphics[width=\linewidth]{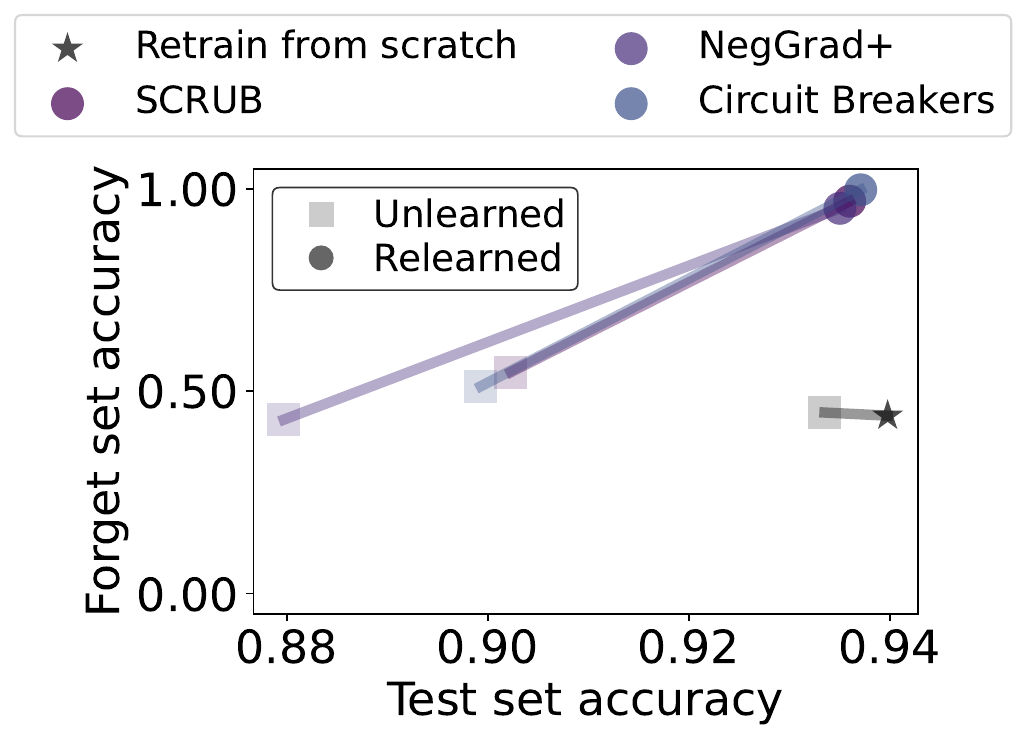}
    \vspace{-6mm}
    \caption{
 \textbf{Fine-tuning an unlearned model on just the retain set recovers performance on the forget set!}
 Results on CIFAR-10 using a forget set of atypical examples from class `airplane'.}%
    \vspace{-6mm}
    \label{fig:relearning_overview}
\end{wrapfigure}

Machine unlearning is the problem of removing the influence of specific training datapoints, the \textit{forget set}, from a pretrained model. This was initially motivated from the perspective of privacy, the right-to-be-forgotten~\citep{cao2015towards,neel2021descent}, data protection policies~\citep{mantelero2013eu}, and recently applied to a range of problems, including removing harmful knowledge~\citep{li2024wmdp,zou2024circuitbreakers,eldan2023whosharry}.

Exact unlearning refers to completely eliminating the influence of the forget set. This objective can be achieved by retraining the model without those datapoints (i.e., \textit{retrain from scratch})~\citep{bourtoule2021shardedtrain}.
However, such a method is computationally prohibitive as it requires retraining the model for every unlearning request~\citep{bourtoule2021shardedtrain}.
This issue motivated the development of approximate unlearning methods, where the aim instead is to only approximately remove the influence of the given datapoints~\citep{chundawat2023badteacher,foster2024ssd,kurmanji2024scrub,li2024wmdp,zou2024circuitbreakers}, in exchange for greater efficiency.
While such methods may seemingly succeed at matching the retrained-from-scratch model on some simple metrics, like the accuracy on the forget set, it is unclear if they permanently remove the influence of these datapoints~\citep{hayes2024inexact}. In fact, evaluating whether they do is a research problem in its own right~\citep{hayes2024inexact,triantafillou2024unlearningcomp}. 

In this work, we take a different perspective, focusing on relearning attacks on unlearned models. We consider \textit{tampering attacks}~\citep{che2025tamperingattacks,tamirisa2408tamper}, where the attacker is able to fine-tune the model weights. This direction is inspired by a growing set of observations in the context of unlearning ``knowledge'' in LLMs: fine-tuning an unlearned model even on seemingly benign data may cause the believed-to-be-unlearned knowledge to re-emerge~\citep{hu2024jogging,lucki2024adversarial,deeb2024unlearning,che2025tamperingattacks}. However, these studies are carried out under conditions that make it difficult to draw clear conclusions. 
First, these works study unlearning \textit{knowledge, capabilities, or topics}, where the problem is inherently under specified. For example, given a dataset containing knowledge necessary for making bioweapons, the goal may be to fully remove the capability of constructing bioweapons, while preserving general knowledge of biology. In this setting, it is hard to draw a clear line between forbidden and permissible knowledge and pinpoint all training examples responsible for acquiring different types of knowledge. Furthermore, knowledge is hard to measure, especially due to nuances of natural language, requiring question-answer evaluation which may be sensitive to the particular phrasing. 
Finally, because LLMs can make complex inferences beyond the training set, not all knowledge that is extractable should be attributed to a failure of unlearning~\citep{shumailov2024ununlearning}. In many cases, the acquisition of knowledge is natural and unavoidable, hence, making it difficult to distinguish between the two in these problem settings.

To address these issues, we study relearning attacks in a setting that allows for controlled experimentation: unlearning specific training examples from (small) vision classification models, a problem for which a plethora of approximate unlearning methods have been developed and tested~\citep{golatkar2021mixed,graves2021amnesiac,goel2022towards,kurmanji2024scrub,foster2024ssd,torkzadehmahani2024improved,zhao2024what,sepahvand2025selective}. In example-level unlearning, the gold standard is clear, namely, to retrain the model without the forget set. As our models are small enough, we can compute the gold-standard solution and use it as the reference point for comparison. Because we consider classification, we can also use accuracy as a simple and well-understood measure of performance. These properties combined allow us to compare the tamper resistance of different unlearning algorithms with the correct reference point, and therefore draw conclusions about their quality.

We evaluate a range of increasingly-complex unlearning algorithms in this setting and discover a surprising finding: for numerous unlearning algorithms, the accuracy of the forget set jumps from around 50\% post-unlearning to nearly 100\% after fine-tuning the unlearned models on \textit{only the retain set}, which is disjoint from the forget set.
\cref{fig:relearning_overview} shows this phenomenon on CIFAR-10 using ResNet-18, after having attempted to unlearn a subset of atypical instances of class `airplane'. We note that a model retrained from scratch without the forget set does not exhibit this behaviour, with the accuracy remaining at 50\%. Therefore, the recovery of forget set accuracy can be safely interpreted as a failure of these algorithms to fully remove the influence of the datapoints in the forget set.  

Our extensive analyses leads to multiple insights. First, unlearning and relearning of typical examples is trivial, and unlearning methods behave similarly to the gold standard. However, we see a stark contrast in their respective patterns of behaviour on a forget set  of atypical examples. Furthermore, taking a weight-space view~\citep{frankle2020linear}, we discover a key characteristic of unlearning algorithms that are better at resisting these attacks: they yield an unlearned model that is distant from the pretrained model in the weight-space. Based on this insight, we propose a new class of unlearning algorithms that are superior in terms of resisting relearning attacks by incorporating terms in their objective that encourage the unlearned model to move far away from the pretrained model in the weight-space.

To summarize, we make the following contributions in this work:
\begin{itemize}
    \item We show that unlearning algorithms fail to \textit{delete} the influence of the forget set, which stays \textit{dormant} and can resurface by fine-tuning \textit{even on just the retain set}.
    \item We identify a key characteristic of methods that are more robust against relearning attacks, namely: the unlearned model is distant from the pretrained model in weight space.
    \item Leveraging this insight, we propose a new class of unlearning methods that attempt to push the unlearned model far away from the pretrained model. These methods are significantly more robust against relearning attacks in comparison to unlearning methods that operate only at the output-level~\citep{kurmanji2024scrub} or the representation-level~\citep{li2024wmdp,zou2024circuitbreakers}.
\end{itemize}

\section{Background and Related Work}

\textbf{Unlearning.}
The problem of machine unlearning was introduced by~\citep{cao2015towards}. The goal is to remove the influence of a ``forget set'' from a model that was trained on a dataset including that set. This was motivated by privacy and right-to-be-forgotten policies~\citep{mantelero2013eu}.
The perfect unlearning method, from the perspective of fully erasing the influence of the forget set, is to simply retrain the model excluding that set. However, the computationally prohibitive training costs make such an approach infeasible in most practical cases. \cite{bourtoule2021shardedtrain} propose to shard the dataset, and train an ensemble model over it, allowing to selectively retrain only the affected parameters. However, the computational cost is still prohibitive in the worst case, while also leading to poorer performance in some cases due to the use of specialized architectures.
These issues motivated the development of approximate methods that accept imperfect unlearning in exchange for greater efficiency. This is the family of methods we focus on in this work. The goal in approximate unlearning is to post-process the trained model as efficiently as possible in order to closely match the model which is retrained from scratch using only a small amount of model fine-tuning~\citep{jia2023l1sparse,chundawat2023badteacher,foster2024ssd,kurmanji2024scrub,li2024wmdp,zou2024circuitbreakers,zhao2024what,triantafillou2024unlearningcomp}. This is a challenging problem as imperfect attempts to erase the influence of the forget set post-hoc may have a number of unwanted side-effects, such as harming the overall utility of the model~\citep{triantafillou2024unlearningcomp}.

\textbf{Unlearning quality metrics.} Since most approximate unlearning methods that are applicable to deep models do not come with theoretical guarantees about the quality of their approximation, we are required to estimate how well they approximate retraining from scratch empirically. This is a research problem in and of itself, and current rigorous metrics are very computationally expensive~\citep{triantafillou2024unlearningcomp,hayes2024inexact}. Furthermore, unlearning entails fundamental trade-offs, such as between forgetting and maintaining the model's utility. This requires a multifaceted evaluation metric that captures relevant factors aside from forgetting quality. Commonly, in vision classification, the accuracy on the retain set and the accuracy on the test set are used to measure model utility. In similar spirit to our work, \textit{time to relearn}~\citep{golatkar2021mixed} quantifies the strength of unlearning by the number of optimization steps required to reacquire forgotten information by directly fine-tuning on it. We instead show that we can restore forget set accuracy even when fine-tuning on only a subset of it, or solely on the retain set.

\textbf{Re-emergence of attempted-to-be-unlearned knowledge via fine-tuning.}
Recent work in language models showed that believed-to-be-unlearned knowledge can re-emerge by fine-tuning on a small subset of the forget set or even on seemingly-unrelated data~\citep{hu2024jogging,lucki2024adversarial,deeb2024unlearning,che2025tamperingattacks}.  
Relatedly, it has also been shown that fine-tuning a language model on benign inputs can reverse the safety tuning of the model~\citep{qi2023finetuning,lermen2023lora}.
A key distinction sets our work apart from all prior efforts.
They study unlearning \textit{knowledge, or capabilities}, rather than \textit{specific training examples}. Their goal is to remove unwanted knowledge beyond the specific instances in the forget set, e.g., fully remove a dangerous capability (such as bioweapon construction) after having unlearned on a specific dataset containing related knowledge~\citep{li2024wmdp}. This problem is inherently less well-specified compared to unlearning specific examples where we have a clear definition of the ideal solution, namely, retraining from scratch without the specific examples. In LLMs, measuring knowledge is also nuanced, requiring question-answering tools, for instance, where the success of extracting knowledge may depend on the phrasing~\citep{zou2023universal}. We study relearning attacks for example-level unlearning in vision classifiers, a setting where the forget set is well-specified and the goal is well-defined and simple to measure.

\section{Problem Formulation}

Let $\mathcal{D}_{tr}$ denote a training set and $\mathcal{A}$ a learning algorithm. Let $\mathcal{M}_{P} = \mathcal{A}(\mathcal{M}_{I}, \mathcal{D}_{tr})$ denote the ``pretrained model'', obtained by training on $\mathcal{D}_{tr}$, starting from a random initialization $\mathcal{M}_{I}$. Now, let $\mathcal{D}_{F} \subset \mathcal{D}_{tr}$ denote a forget set that we want to unlearn, and let $\mathcal{D}_{R} = \mathcal{D}_{tr} \setminus \mathcal{D}_{F}$ denote the retain set.

The goal of an unlearning algorithm $\mathcal{U}$, is to post-process the pretrained model $\mathcal{M}_{P}$ to remove the influence of $\mathcal{D}_{F}$. Specifically, we denote an unlearning algorithm by $\mathcal{U}: \mathcal{M} \times \mathcal{D}_R \times \mathcal{D}_F \mapsto \mathcal{M}$ that takes in a model, retain set $\mathcal{D}_R$, forget set $\mathcal{D}_F$, and returns an unlearned model $\mathcal{M}_{U} = \mathcal{U}(\mathcal{M}_{P}, \mathcal{D}_{R}, \mathcal{D}_{F})$. Ideally, the unlearned model $\mathcal{M}_{U}$ should match the gold-standard ``retrained-from-scratch'' model  $\mathcal{M}_{RS} = \mathcal{A}(\mathcal{M}_{I}, \mathcal{D}_{R})$ which starts from a random initialization and trains on only the retain set, fully eliminating the influence of $\mathcal{D}_{F}$. We desire unlearning algorithms that can approximate that solution but are much more efficient than retraining from scratch.

In this work, we study relearning attacks that apply a further fine-tuning phase attempting to reintroduce the forget set. Such attacks that are able to modify the model's weights are referred to in the literature as \emph{tampering attacks}~\citep{che2025tamperingattacks,tamirisa2408tamper}. We carry out these attacks by fine-tuning the model on the union of $\mathcal{D}_{R}$ and a subset of ``relearning examples'' $\mathcal{D}_{F_{re}} \subset \mathcal{D}_{F}$. We denote the relearned model as $\mathcal{M}_{RL} = \mathcal{A}'(\mathcal{M}, \mathcal{D}_{R} \cup \mathcal{D}_{F_{re}})$, where $\mathcal{A}'$ denotes a fine-tuning algorithm used for relearning (which might be similar to $\mathcal{A}$ with slightly different hyperparameters) and $\mathcal{M}$ can be either $\mathcal{M}_{U}$ or $\mathcal{M}_{RS}$. 

We measure performance on the held-out portion of the forget set (held-out from the perspective that it was not used during relearning), denoted $\mathcal{D}_{F_{ho}} = \mathcal{D}_{F} \setminus \mathcal{D}_{F_{re}}$. We vary the size of $\mathcal{D}_{F_{re}}$ and measure the effect on relearning. 
We also measure performance on a test set $\mathcal{D}_{te}$, to measure utility. 

An ideal unlearning algorithm is one that is \textit{tamper resistant}: upon relearning, its accuracy on the forget set does not increase more than it would by learning the relearning set anew. In other words, the forget set accuracy of $\mathcal{A}'(\mathcal{M}_U, \mathcal{D}_{R} \cup \mathcal{D}_{F_{re}})$ should not be higher than that of $\mathcal{A}'(\mathcal{M}_{RS}, \mathcal{D}_{R} \cup \mathcal{D}_{F_{re}})$. At the same time, an ideal unlearning algorithm would not sacrifice the test accuracy.

\textbf{Threat model.}
Similar to tampering attacks considered for LLMs~\citep{tamirisa2408tamper,lucki2024adversarial,hu2024jogging,deeb2024unlearning,che2025tamperingattacks}, we assume that the defender has access to a pretrained model, and performs unlearning using any algorithm of their choice. Furthermore, we assume that the attacker has white-box access to the unlearned model provided by the defender, the retain set $\mathcal{D}_{R}$, and limited access to the forget set (i.e., the relearning set $\mathcal{D}_{F_{re}}$). The goal of the attacker is to recover performance on the full forget set $\mathcal{D}_{F}$, while minimizing the number of unlearned examples needed $\mathcal{D}_{F_{re}}$ (as relearning becomes trivial if $\mathcal{D}_{F_{re}} = \mathcal{D}_{F}$). We also consider an extreme---and perhaps more realistic---case of access to only $\mathcal{D}_{R}$.

\section{Experimental Setup}
\label{sec:exp_setup}

\textbf{Models and Datasets.}
We use two different models for our evaluation from the ResNet model family~\citep{he2016deep}, namely ResNet-18 and ResNet-34~\citep{he2016deep}.
In terms of datasets, we use CIFAR-10 and CIFAR-100 datasets~\citep{krizhevsky2009learning} with 10 and 100 classes respectively, and a total of $50k$ training instances in each case ($5k$ instances per class for CIFAR-10, and $500$ instances per class for CIFAR-100).

\textbf{Evaluation.}
All models are evaluated in terms of accuracy on the held-out part of the forget set  $\mathcal{D}_{F_{ho}}$ (same subset between all models), as well as the test set $\mathcal{D}_{te}$.
While we can report accuracy on the full forget set $\mathcal{D}_{F}$ for the unlearned model, we instead report accuracy on the remaining subset in order to enable a direct comparison of the impact of the relearning attack.
The line plots show accuracy every 10 optimization steps. 
When reporting results using a scatter plot, we average the test set accuracy as well as the accuracy on the forget set for the last 50 steps reported in the line plots.

\textbf{Pretraining.}
We pretrain the model for 300 epochs using Adam optimizer~\citep{kingma2014adam} with a learning rate of 1e-4, cosine learning rate decay with a decay factor of 0.1, batch size of 128, and a weight decay of 1e-4 in all configurations.

\textbf{Unlearning.}
We consider two unlearning settings: \emph{sub-class unlearning}, where the forget set consists of 10\% of the class instances (here, sub-class means a subset of the complete class), and \emph{class-agnostic unlearning}, where we select 1\% of the data set regardless of class labels.
This ensures that we use the same number of examples in the forget set for both settings on CIFAR-10 (we only evaluate sub-class unlearning on CIFAR-100).
We use a smaller learning rate of 1e-5 without any weight decay and optimize the model for a 100 epochs during the unlearning phase.

\textbf{Relearning attack.}
During this phase, we fine-tune on a combination of the retain set $\mathcal{D}_{R}$ and a subset of the forget set for relearning ($\mathcal{D}_{F_{re}}$). We explore the impact of different choices for relearning examples in \cref{app:relearn_ex_type}. 
We again use a small learning rate of 1e-5 without any weight decay, and optimize the model for just 10 epochs (except \cref{fig:relearning_overview} where we optimized the model for 300 epochs).
Similar to the pretraining stage, we use a cosine learning rate decay with a decay factor of 0.1.

\subsection{Baseline Unlearning Methods}

\begin{figure}[t]
    \centering
    \includegraphics[trim={0 0 0 0},clip,width=0.98\linewidth]{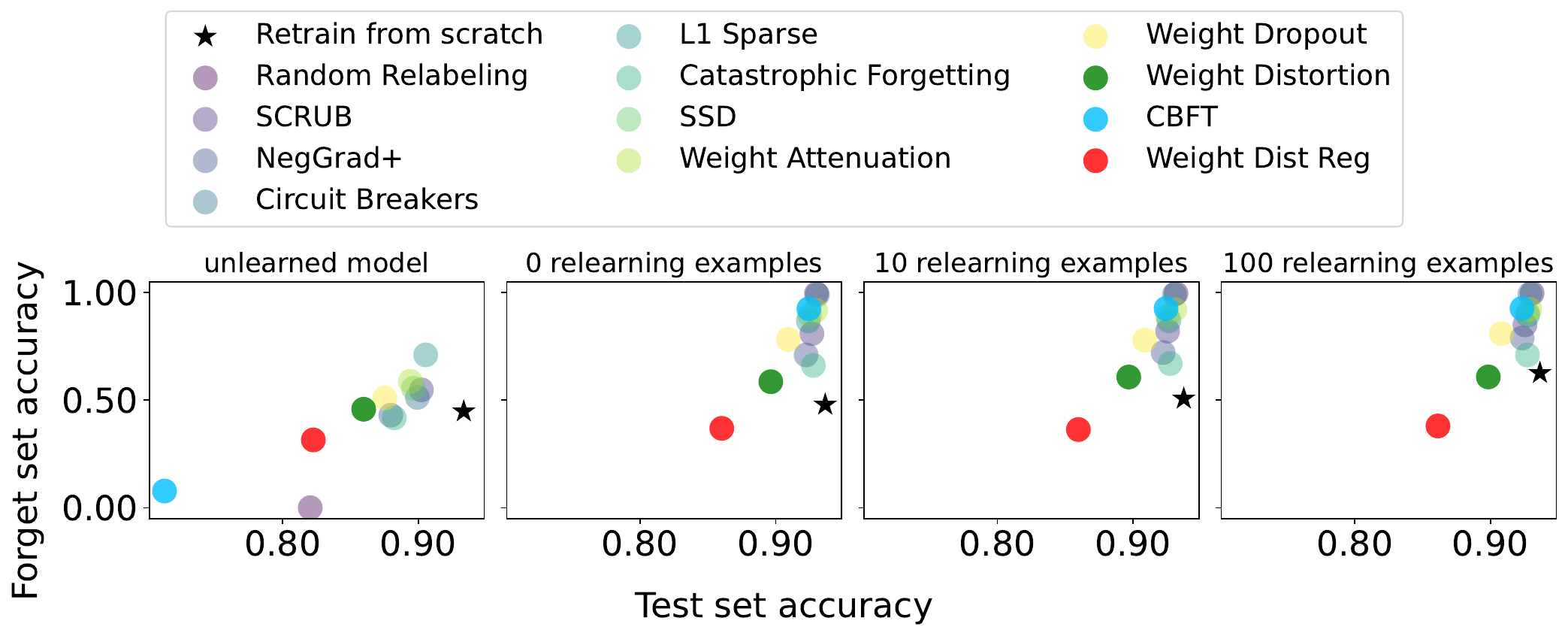}
    \includegraphics[trim={0 0 0 55mm},clip,width=0.98\linewidth]{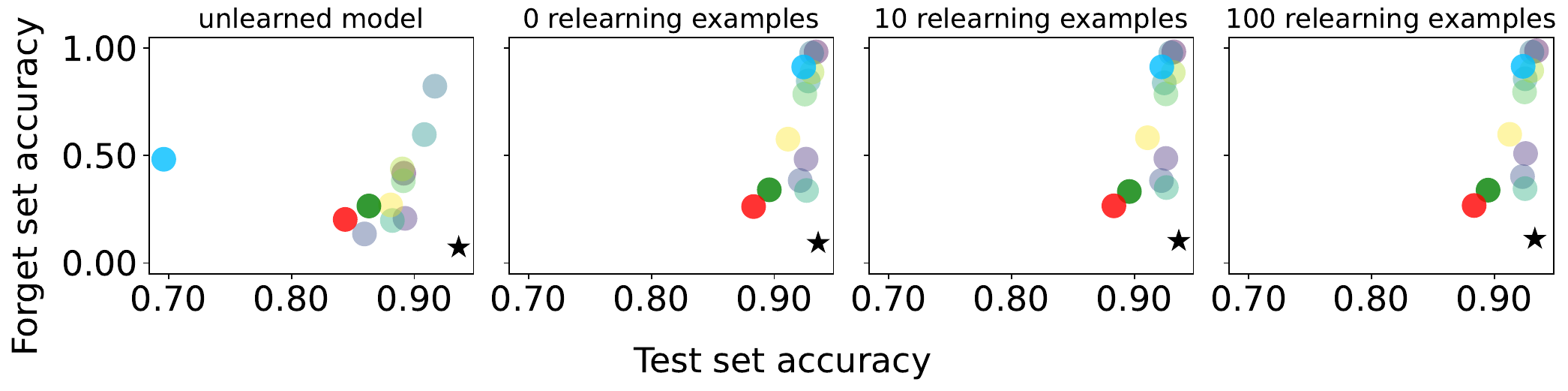}
    \vspace{-5mm}
    \caption{
    Scatter-plots with test-set accuracy on the x-axis and accuracy on the held-out portion of the forget set,
    $\mathcal{D}_{F_{ho}}$, on the y-axis.
    The left-most subplot indicates performance immediately following unlearning. The next three subplots are following a relearning attack with instances of the retain set, $\mathcal{D}_R$ and a varying number of instances of the forget set, $\mathcal{D}_{F_{re}}$ (0, 10, and 100, respectively).
    Each point is the average performance of the last 50 steps (see \cref{fig:relearning_forget_cifar10_resnet18_high_mem_evolution} for the whole trajectory for sub-class unlearning and \cref{fig:relearning_forget_cifar10_resnet18_high_mem_all_cls_remaining} for the trajectory for class-agnostic unlearning). The forget set is comprised of atypical examples (from the `airplane' class, i.e., sub-class unlearning for the top row and all classes, i.e., class-agnostic unlearning in the bottom row) in CIFAR-10.
    The figure indicates that many methods achieve near-perfect recovery of unlearned knowledge with only a small amount of model fine-tuning, even with 0 relearning examples (fine-tuning on only the retain set). Weight Distortion, CBFT, and Weight Dist Reg are introduced in \cref{sec:weight_space_analysis}.}
    \vspace{-3mm}
    \label{fig:relearning_forget_cifar10_resnet18_high_mem}
\end{figure}

We consider a range of different baseline unlearning methods.
Each method has its own set of hyperparameters. We attempted to select the hyperparameters to achieve a good trade-off between the test accuracy and the forget set accuracy for each unlearning method.

\textbf{SCRUB}~\citep{kurmanji2024scrub} uses a two-phase training procedure where it interleaves iterations on the forget set and the retain set. The loss function minimizes KL-divergence between the pretrained model and the unlearned model output distributions on the retain set, along with cross-entropy on the true labels.
For unlearning, it maximizes the KL-divergence on the forget set between the distributions of the pretrained model and the unlearned model.

\textbf{Circuit Breakers}~\citep{zou2024circuitbreakers,li2024wmdp} was proposed particularly to unlearn knowledge in language models. The training procedure attempts to push the representations apart by minimizing cosine similarity with the pretrained model on the forget set, while minimizing the Euclidean distance of the representations on the retain set to avoid model collapse. We apply circuit breaker loss on layer 4 and layer 7 of our models, which is motivated by the fractional depth considered in the original work.

\textbf{NegGrad+}  ~\citep{kurmanji2024scrub} maximizes the cross-entropy loss on the forget set, while minimizing the loss on the retain set. We used the alternating variant (similar to SCRUB) instead of joint optimization of the two losses, as it resulted in better test accuracy as well as lower forget set accuracy.

\textbf{Catastrophic Forgetting}~\citep{triantafillou2024unlearningcomp} uses repeated fine-tuning on the retain set with a weight decay, which naturally leads to a decay in the magnitude of the parameters that are unimportant for the forget set. We use a weight decay of 0.001 for all our models.

\textbf{L1-Sparse}~\citep{jia2023l1sparse} is similar to the Catastrophic forgetting in our case, except that it minimizes $L_1$-norm instead of the $L_2$-norm employed in weight decay.

\textbf{Selective Synaptic Dampening (SSD)}~\citep{foster2024ssd}  identifies model weights to dampen based on their importance for the retain set and forget set, quantified using the Fisher information matrix. In contrast to the original paper, we follow this process with fine-tuning on the retain set to be consistent with our other baselines, giving SSD a better shot at repairing test accuracy.

\textbf{Random Relabeling}~\citep{graves2021amnesiac} relabels every example in the forget set with a random label. Note that we used an aggressive version of random relabeling where we re-assign a new label at every fine-tuning step. Hence, this can be considered analogous to minimizing divergence to a uniform distribution.

\textbf{Weight Attenuation}  attenuates all model weights with a fixed attention factor of 0.5, followed by simple fine-tuning on the retain set.

\textbf{Weight Dropout} performs random (unstructured) pruning with a dropout factor of 0.2 (i.e., zeroing out 20\% of the model weights), followed by simple fine-tuning on the retain set.

\textbf{Tampering Attack Resistance (TAR)} ~\citep{tamirisa2408tamper}, originally proposed for LLM unlearning, defines a bi-level optimization, starting from an already unlearned model, with the aim to make it resistant against tampering attacks. It uses a first-order approximation of inner adversaries, and attempts to maximize the entropy of the model predictions after fine-tuning of the unlearned model. TAR also uses a representation alignment loss which minimizes the Euclidean distance between the representations of the initially unlearned model and the current model to avoid model collapse. Similar to Circuit Breakers, we apply the representation alignment loss on layer 4 and layer 7 of our models. Note that TAR relies on an unlearned model as a starting point, making it a two-phase approach.

\section{Recent Unlearning Algorithms are not Tamper-Resistant}
\label{sec:main_results}

\begin{figure}[t]
    \centering
    \includegraphics[width=0.8\linewidth]{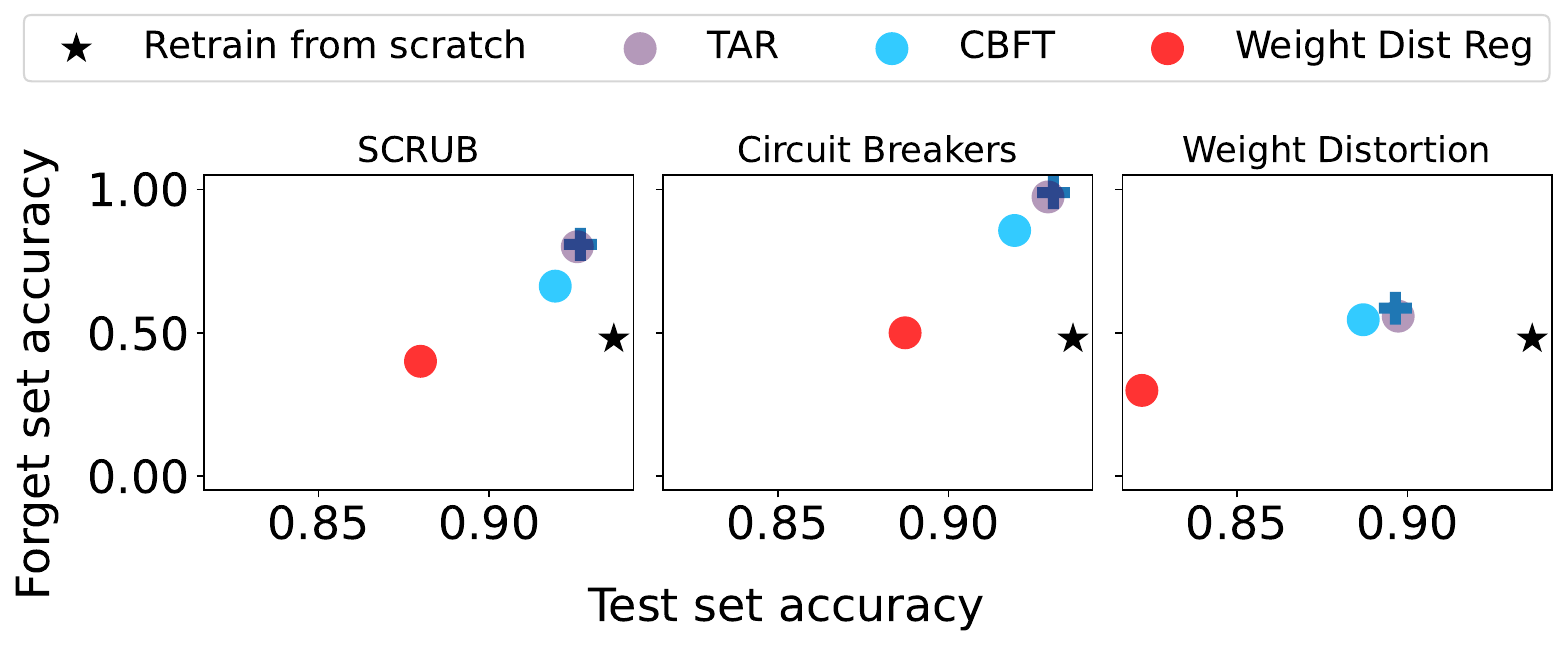}
    \vspace{-3mm}
    \caption{
    Comparison between test set accuracy and accuracy on the held-out part of the forget set $\mathcal{D}_{F_{ho}}$ after relearning, for subclass unlearning of atypical examples in CIFAR-10. We consider two-phase unlearning methods: first, an initial safeguard (unlearning phase) is applied, with the unlearning algorithm mentioned as the subplot title. Then, each of TAR, CBFT, and Weight Dist Reg are applied as a second phase for increasing the tamper-resistance. The `+' symbol represents the performance of the initial safeguard for reference. We observe that TAR fails to add any tamper-resistance in addition to that of the initial safeguard despite being designed for this.}
    \label{fig:relearning_forget_cifar10_resnet18_high_mem_new_init_sg}
\end{figure}

We begin by investigating the tamper-resistance of current state-of-the-art unlearning methods. 
We use CIFAR-10 and CIFAR-100 with ResNet-18, and attempt to unlearn instances of a single class (`airplane' class for CIFAR-10, and `apple' class for CIFAR-100), or across classes (\emph{class agnostic}).

\textbf{The role of typicality for learning and unlearning.}
Not all examples are equally easy to unlearn, and not all examples are equally easy to (re)learn. Typical examples can be particularly easy to unlearn as even a no-op unlearning already yields similar predictions on these examples as the retrained from scratch model~\citep{jia2023l1sparse}. This is because they are, by definition, easy to predict regardless of whether they are part of the training set or not~\citep{feldman2020neural,jiang2020characterizing,baldock2021deep,siddiqui2022metadata}. Generally, because typical examples are more likely to be predicted correctly, they are easier to relearn as well. Based on this, we hypothesize that the typicality of examples is a key property that will determine relearning patterns and differences between unlearning algorithms compared to retraining from scratch.

To investigate this, we study three different settings, characterized by different typicality levels, where the forget set contains instances of the `airplane' class that are: (i) most likely to be typical, (ii) randomly selected, and (iii) most likely to be atypical. We leverage pre-computed consistency scores from \citep{jiang2020characterizing} to separate typical and atypical instances, by treating instances with the highest consistency scores as typical instances, while treating instances with the lowest consistency scores as atypical instances.
Other scoring schemes are also equally applicable ~\citep{feldman2020neural,siddiqui2022metadata,baldock2021deep}.

When the forget set consists of typical examples, relearning is trivial and uninteresting. The reason is that accuracy of the retrain-from-scratch model is essentially perfect on the forget set, even before relearning is applied, and all unlearning methods exhibit similar behavior (for the sake of completeness, we show the results for this case in \cref{fig:relearning_forget_cifar10_resnet18_low_mem} -- \cref{app:typical_ex}).
Similarly, when the forget set consists of randomly selected examples, performance of retrain-from-scratch is nearly perfect because randomly selected examples are predominantly typical~\citep{feldman2020neural,feldman2020does,baldock2021deep,siddiqui2022metadata} (again, for completeness, this case is shown in \cref{fig:relearning_forget_cifar10_resnet18_random} -- \cref{app:random_ex}). Consequently, we focus on atypical forget set items for the remainder of the paper; in this case, retrain-from-scratch will not predict correctly prior to relearning and thus we can measure the effectiveness of a relearning attack.

\textbf{Relearning attacks succeed against several unlearning baselines}. 
We present the results for sub-class unlearning for the forget set of atypical examples in \cref{fig:relearning_forget_cifar10_resnet18_high_mem} (top).
In this and subsequent figures, existing methods are indicated by pastel colors. New methods, to be introduced shortly, are represented by saturated colors. For the most part, the existing methods all behave similarly and the reader need not attend to the individual methods.
The accuracy of \textit{retrain from scratch} is less than 50\% on that forget set, and remains almost exactly the same when subjected to the relearning attack of fine-tuning only on the retain set (i.e., 0 relearning examples).
The accuracy of this model shifts up slightly as we increase the number of relearning examples from the forget set (going from left to right).
In stark contrast, the different unlearning methods we evaluate show a qualitatively different trend. We make the striking observation that some methods (such as Circuit Breakers, SCRUB, and Random Relabeling) are very susceptible to relearning attacks. For these methods, forget-set accuracy drops down after unlearning, near the desired reference point of the retrained model. However, upon relearning even on just the retain set, the model achieves near-perfect accuracy on the forget set---a jump from near 50\% post-unlearning to nearly 100\% after relearning.

\textbf{Sub-class vs class-agnostic unlearning}.
We further attempt to understand if such differences exist in a class-agnostic forget-set setting (with the same number of forget-set examples, i.e., 500), which is now comprised of atypical examples across classes.
\cref{fig:relearning_forget_cifar10_resnet18_high_mem} (bottom) shows that class-agnostic unlearning better differentiates among methods compared to sub-class unlearning. We see a wider spread among methods, even in the first subplot (post-unlearning, pre-relearning), since class-agnostic unlearning is harder. We also observe that while for sub-class unlearning, the performance of retrain-from-scratch (black star) on $\mathcal{D}_{F_{ho}}$ shifts upwards as more relearning examples are used, this shift does not occur with class-agnostic unlearning. This result is expected since relearning on a larger number of atypical examples does not improve performance on a disjoint set of other atypical examples (by definition of atypicality~\citep{feldman2020neural}). There is again a stark contrast between the performance of retrain-from-scratch model on $\mathcal{D}_{F_{ho}}$, where post-relearning accuracy remains very low, and many unlearning algorithms, where post-relearning accuracy again reaches nearly 100\%.
Importantly, the relative ranking between methods in the sub-class and the class-agnostic case remains consistent.

\textbf{Relearning attacks succeed even against methods designed for tamper-resistance}. We further compare two-phase methods that assume an initial unlearning round, followed by a subsequent training round in order to reduce susceptibility to relearning attacks. This is inspired by the methodology of TAR~\citep{tamirisa2408tamper}, which is explicitly designed to increase the resistance of the unlearned model against fine-tuning-based relearning attacks. 
Despite TAR's success in the case of language models, \cref{fig:relearning_forget_cifar10_resnet18_high_mem_new_init_sg} shows that it struggles to provide any resistance against relearning attacks in our case. 

\section{A Weight-Space View on Understanding and Improving Tamper-Resistance}
\label{sec:weight_space_analysis}

\begin{figure}[t]
    \centering
    \includegraphics[width=\linewidth]{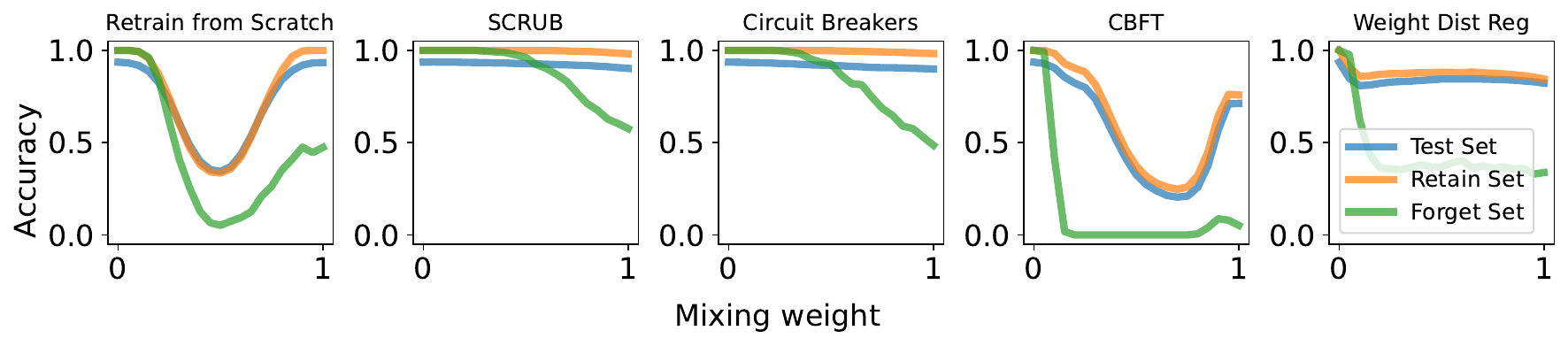}
    \vspace{-6mm}
    \caption{Linear mode connectivity analysis on CIFAR-10, where the forget set is comprised of atypical examples.
    We construct a linear path between the pretrained and the unlearned (or retrained-from-scratch) model by interpolating the model parameters and batch-norm statistics using different mixing weights (shown on the x-axis). We report accuracy on the y-axis. 0 on the x-axis represents the pretrained model, while 1 represents the unlearned or retrained model. Retrain-from-scratch is not linearly connected to the pretrained model, whereas for unlearning algorithms, the resulting unlearned model is in many cases still linearly connected to the pretrained one.}
    \label{fig:linear_mode_conn_cifar10_resnet18_high_mem}
\end{figure}

In the previous section, we demonstrated the susceptibility of existing prominent unlearning methods to tampering (\cref{fig:relearning_forget_cifar10_resnet18_high_mem}). However, it is unclear what makes a method vulnerable or robust to these relearning attacks. In this section, we shed light on this  question through a weight-space view. Specifically, we hypothesize that the susceptibility of an unlearned model to relearning may be associated with failing to move `far enough' from the pretrained model in weight-space. We explore this hypothesis from two perspectives: (i) by measuring distances in weight-space, and (ii) through Linear Mode Connectivity analysis~\citep{frankle2020linear}. We then use these tools to interpret the tamper-resistance profiles of previously-proposed unlearning algorithms (\cref{fig:relearning_forget_cifar10_resnet18_high_mem}) that show better tamper-resistance (such as Catastrophic Forgetting and L1 Sparse) compared to the ones that exhibit worse tamper-resistance (such as Random Relabeling, Circuit Breakers, and SCRUB).

\begin{wrapfigure}{r}{0.5\textwidth}
    \centering
    \centering
    \includegraphics[width=\linewidth]{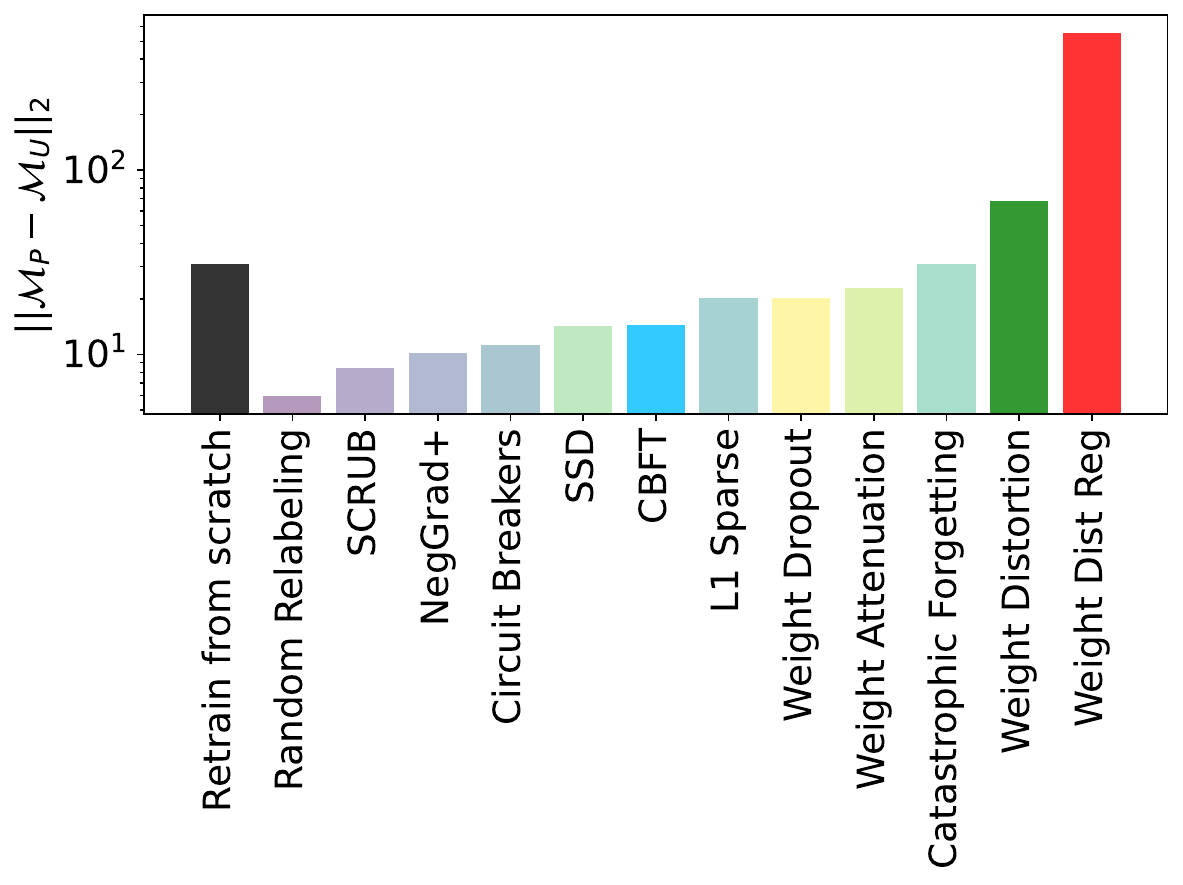}
    \vspace{-5mm}
    \caption{$L_2$ norm of the difference between the parameters of the pretrained and the unlearned models induced by different unlearning methods. We consider only the difference in the parameters, while ignoring the batch-norm statistics for ResNet-18 trained on CIFAR-10, where the forget set is comprised of atypical examples.}
    
    \vspace{-3mm}
    \label{fig:param_diff_cifar10_resnet18_high_mem}
\end{wrapfigure}

\paragraph{Weight-space distance.}
\cref{fig:param_diff_cifar10_resnet18_high_mem} plots $L_2$ distance between the pretrained model and the unlearned model.
We see that existing methods that we previously showed to be susceptible to relearning (pastel colors)  induce only small movement in parameter space, indicated by the small $L_2$ distance. 

Generally, we see that methods with higher $L_2$ distance exhibit higher robustness against relearning attacks presented in \cref{fig:relearning_forget_cifar10_resnet18_high_mem}. Notably, two methods based on these insights 
which we describe shortly, \emph{Weight Distortion} and \emph{Weight Dist Reg},
have the highest $L_2$-norm and higher tamper-resistance. We remark that, out of previous methods, those with increased tamper-resistance (such as Catastrophic Forgetting and L1 Sparse) have comparatively higher distance norm compared to methods with very poor tamper resistance (such as Random Relabeling, Circuit Breakers, and SCRUB). We note that both Catastrophic Forgetting and L1 Sparse use  weight-space regularizers ($L_2$ and $L_1$, respectively). Fine-tuning without any regularizer failed to unlearn the forget set in our evaluations, which aligns with our weight-space interpretation.
We consider more elaborate analysis regarding layers that are particularly important for robustness against relearning attacks as an interesting direction for future work.

\paragraph{Linear Mode Connectivity (LMC).}

We further investigate the relationship in weight-space between the unlearned and the retrained from scratch model via the lens of LMC~\citep{frankle2020linear}.
We plot the accuracy when interpolating between the two models (where we interpolate both model parameters as well as model batch-norm statistics) along a linear path.
We compare the pretrained and retrained from scratch model (which are trained from the same initialization), as well as the pretrained and different unlearned models in \cref{fig:linear_mode_conn_cifar10_resnet18_high_mem}.
Looking at the retrain from scratch plot, we see that there is a high-loss barrier between the two models, meaning that they are not in linearly connected modes. The same holds for methods with some tamper resistance (like Catastrophic Forgetting and L1 Sparse) but not for those vulnerable to tampering attacks like SCRUB, where we observe no such barrier.

\subsection{A New Class of Tamper-Resistant Unlearning Methods}
\label{sec:new_algos}

We now leverage these insights to propose a class of unlearning methods that are designed with tamper-resistance in mind. We achieve this through objectives that either aim to induce a large distance in the weight-space, or a loss barrier between the pretrained and unlearned models.
Hence, any method that directly or indirectly attempts to separate out the pretrained model and the unlearned model is an instantiation of this framework.

\textbf{Weight Distortion.} This is a very simple method that adds isotropic zero-mean Gaussian noise to all model weights with a fixed standard deviation of 0.02, followed by simple fine-tuning on the retain set. We hypothesize that the addition of noise facilitates moving away from the pretrained model.

\textbf{Weight Dist Reg.} We directly aim to maximize the distance between the pretrained and unlearned models, by explicitly adding a term that quantifies the Euclidean distance between the two models. We maximize this loss during training, while minimizing the loss on the retain set.

\textbf{Connectivity-Based Fine-Tuning (CBFT).}
We employ the method from \citep{lubana2023mechanistic}, which was originally proposed to obtain models focusing on distinct recognition mechanisms.
We  maximize the loss on the midpoint between the pretrained model and the current unlearned model on examples from the retain set as well as the forget set, while only minimizing the loss on the retain set for the unlearned model.
This attempts to add a high-loss barrier in between, while still retaining model utility on the retain set for the final unlearned model.
We use a small weighting factor of 0.001 on the loss maximization term.
Similar to \citep{lubana2023mechanistic}, we ignore the loss maximization term if the loss magnitude exceeds a value of 50.

\textbf{Findings}. \cref{fig:relearning_forget_cifar10_resnet18_high_mem} shows that Weight Distortion and Weight Dist Reg are significantly more tamper-resistant in comparison to prior approaches.
However, we observe that CBFT is less effective than Weight Dist Reg and Weight Distortion across the board. We hypothesize that this is due to acting on model outputs and only indirectly influencing the weight-space. Indeed, while CBFT does create a larger loss barrier than other methods (\cref{fig:linear_mode_conn_cifar10_resnet18_high_mem}), the $L_2$ norm between the pretrained and unlearned parameters is relatively low (\cref{fig:param_diff_cifar10_resnet18_high_mem}). Overall, out of the two weight-space diagnostic tools, we find that the $L_2$ norm of the difference in model parameters is more reliable in predicting tamper-resistance.
\cref{fig:relearning_forget_cifar10_resnet18_high_mem_new_init_sg} shows that both Weight Dist Reg and CBFT, when applied as a second phase on an initial safeguard, can substantially improve its tamper-resistance, unlike TAR. The only exception is if the initial safeguard is Weight Distortion, which already has sufficient tamper-resistance.

\section{Analysis and Discussion}

\textbf{On the trade-off between tamper-resistance and test accuracy}. As discussed previously, there are inherent trade-offs in unlearning, between forgetting the specified examples while maintaining utility (measured via test accuracy)~\citep{kurmanji2024scrub,deeb2024unlearning,triantafillou2024unlearningcomp}. Here, we discover a different trade-off between resisting relearning attacks, and test accuracy. Indeed, \cref{fig:relearning_forget_cifar10_resnet18_high_mem} shows that the methods that are best at defending against these attacks are the ones with the lowest test accuracy. Having surfaced this fundamental tension, we hope future work improves on the current Pareto frontier formed by our new methods.

\textbf{Findings hold across datasets/architectures}. The results presented in the paper are consistent across models i.e., ResNet-34 on CIFAR-10 as presented in \cref{fig:relearning_forget_cifar10_resnet34_high_mem} -- \cref{app:resnet34}.
Furthermore, these results are also consistent across different datasets, i.e., CIFAR-100, as highlighted in \cref{fig:relearning_forget_cifar100_resnet18_high_mem} -- \cref{app:cifar100}.
Finally, our findings are also consistent on a higher-resolution Imagenette dataset~\citep{imagenette} as highlighted in \cref{tab:imagenette_results} -- \cref{app:imagenette}.

\textbf{Efficiency of Unlearning Algorithms}. We used a fix budget of 100 epochs for unlearning to avoid confounding from differences in optimization efficiency.
In order to justify the use of approximate unlearning algorithms instead of exact unlearning (i.e., \textit{retrain from scratch}) with a training budget comparable to pretraining, we evaluate the efficiency of unlearning algorithms in \cref{tab:unlearning_efficiency} -- \cref{app:unlearning_efficiency}.
The results show that our methods are much more efficient at unlearning (and consequently resisting re-emergence of knowledge) even after a single epoch of unlearning compared to other approaches.

\begin{figure}[t]
    \centering
    \includegraphics[width=\linewidth]{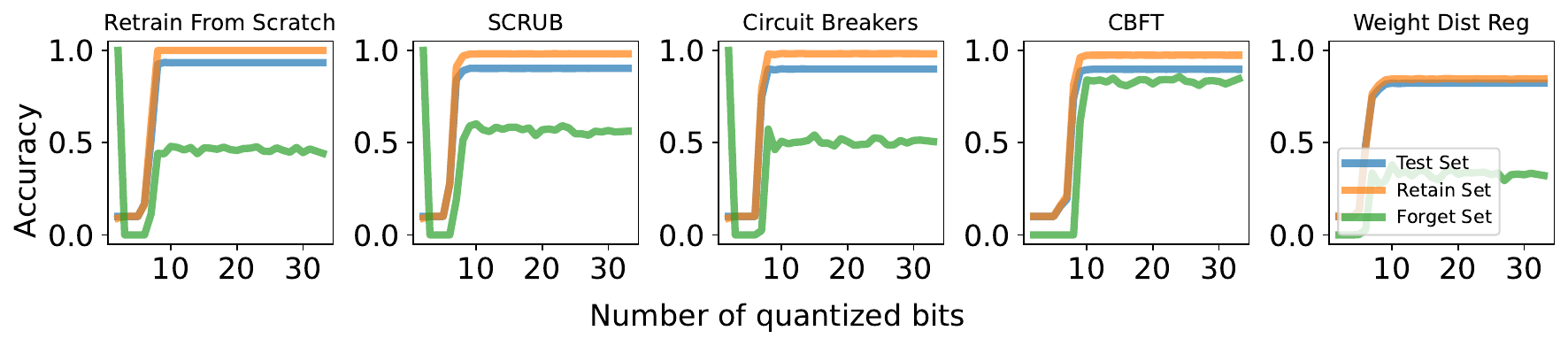}
    \vspace{-7mm}
    \caption{Comparison between the number of quantized bits and accuracy of the model for different unlearning methods on CIFAR-10 and ResNet-18, where the forget set is comprised of atypical examples. In contrast to language models, where quantization has been shown to be responsible for the re-emergence of unlearned knowledge~\citep{zhang2024catastrophic}, we observe that quantization has little to no impact on forget set accuracy until the point where full performance is recovered. Note that the accuracy of 100\% at the start in most cases is an artifact of the model prediction collapsing to just class 0 (whose instances have been selected as our forget set).}
    \label{fig:quantization_attack}
    \vspace{-3mm}
\end{figure}

\textbf{Difference between Quantization and Relearning Attacks}.
Language models have been shown to be susceptible to quantization attacks~\citep{zhang2024catastrophic}, where quantization after unlearning is sufficient to recover unlearned knowledge from the model.
We evaluate susceptibility to quantization in \cref{fig:quantization_attack}.
Interestingly, our vision models are not vulnerable to these quantization attacks: forget set accuracy stays at the same level as the full-precision unlearned model regardless of the number of bits until it gets close to 8 bits, at which point the model collapses. Thus, the spontaneous-recovery phenomenon we explore via relearning attacks is distinct from the recovery obtained via quantization.
We hypothesize that the primary reason why quantization attacks succeed in language models is due to their larger number of parameters (on the order of 100 billion parameters~\citep{dubey2024llama,liu2024deepseek}), resulting in very small changes in weights during unlearning. Hence, models can easily revert to their previous values with quantization.
This problem is further exacerbated with models that are trained with quantization in mind. Such models can immediately latch onto the previous values that were regularized to be easily quantizable~\citep{jung2019learning}.
Similar to our findings, \citep{zhang2024catastrophic} hypothesized that small differences are the cause of this susceptibility to quantization attacks.
They further proposed increasing the learning rate as a potential mechanism to induce larger differences between the pretrained and unlearned models~\citep{zhang2024catastrophic}, while our work introduces more principled approaches for the same purpose.

\textbf{The role of the retain set for relearning}. Given the surprising finding that we can recover forget set accuracy while fine-tuning on only the retain set, we ask: are the retain set examples necessary for this to occur, or could we have used any other examples from the same distribution? We investigate this in  \cref{app:relearn_ex_type} where we replace the retain set with different sets, and fine-tune the unlearned model on those different sets (combined with the `relearning set', which is a subset of the forget set, as before). Notably, replacing the retain set with test examples (that the model was not previously trained on) causes the relearning effect to be a bit less pronounced, especially when 0, or few, relearning examples are used, highlighting the importance of using training rather than held-out data for inducing relearning. This observation relates to other recent findings on anticipatory knowledge reawakening when exposing the model to a repeated sequence of documents. In this scenario, as the model processes documents in a fixed order, it unexpectedly begins to recover an increasing amount of information about a previously seen example even before encountering that example again~\citep{yang2024reawakening}.

\section{Conclusion}

We showed that unlearning methods are susceptible to relearning attacks where the forget set accuracy can be recovered simply by fine-tuning on the retain set. For atypical examples in particular, there is a stark contrast between the relearning patterns of unlearning methods compared to retraining from scratch. Based on weight-space analysis, we suggest two diagnostic tools for understanding tamper-resistance, and propose simple methods that yield state-of-the art tamper-resistant results, revealing new pathways for improving unlearning. The authors of ~\citep{sepahvand2025selective} argued that unlearning algorithms that operate on representations rather than at the level of outputs may be more robust at defending some types of attacks. Our findings take this discussion further, showing that \textit{unlearning methods that operate at either the level of the model outputs or the representations, without any constraint on model weights} (which include methods such as SCRUB, Circuit Breakers, and Gradient Ascent, etc) \textit{should struggle to induce robustness against relearning attacks.}
On the other hand, methods that \textit{directly or indirectly push the pretrained and the unlearned models apart via any intervention} (which includes distorting model weights, regularizing the model to decay parameter magnitude, or even directly pushing the models apart using an explicit loss term) \textit{should be significantly more robust} to relearning attacks. Our proposed methods for increasing tamper-resistance exemplify this and we hope future work builds on these further.

\section*{Acknowledgements}

The authors would like to acknowledge useful discussions with Yanzhi Chen and Ilia Shumailov regarding unlearning, and susceptibility to relearning.
We are also very thankful to the anonymous reviewers for their useful feedback on the initial submission.
AW acknowledges support from Turing AI Fellowship under grant EP/V025279/1, the Alan Turing Institute, and the Leverhulme Trust via CFI.

\bibliography{main}
\bibliographystyle{plain}

\newpage
\FloatBarrier
\appendix

\section{Tamper-Resistance with Typical Examples for the Forget Set}
\label{app:typical_ex}

\begin{figure}[t]
    \centering
    \includegraphics[width=\linewidth]{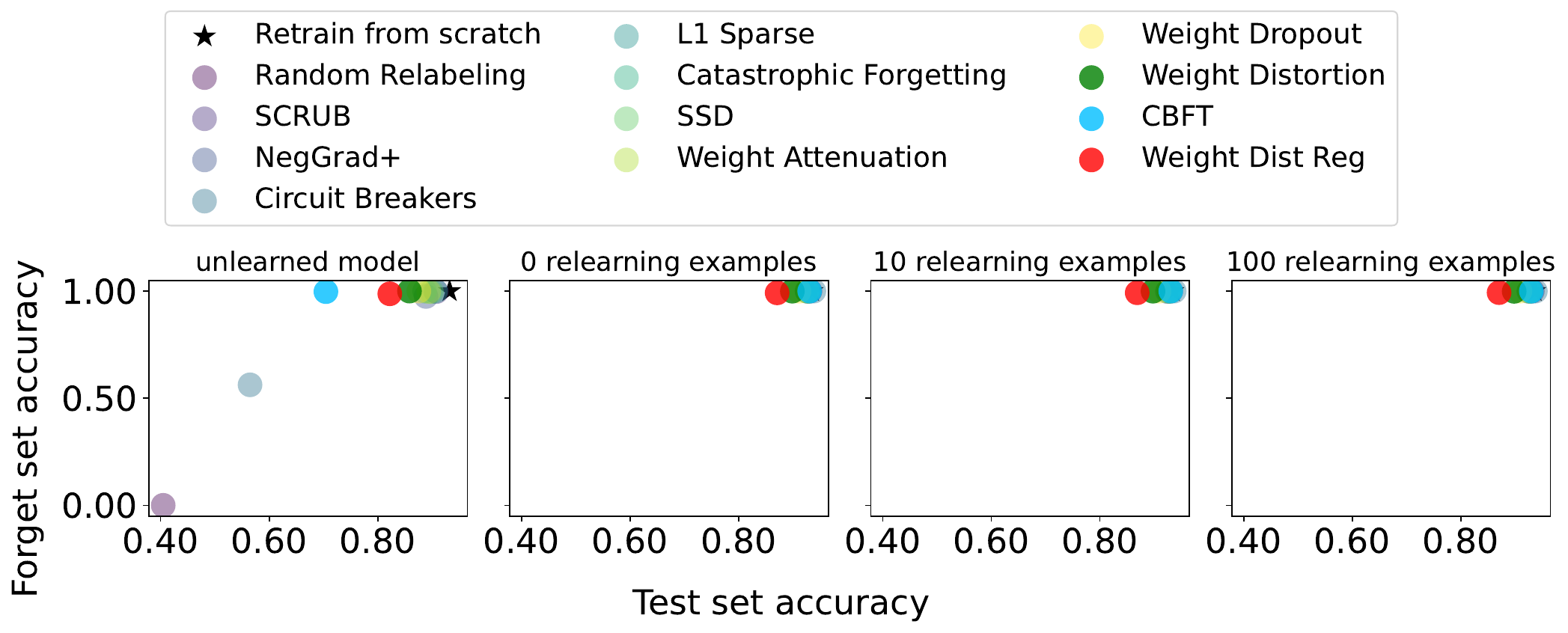}
    \includegraphics[width=\linewidth]{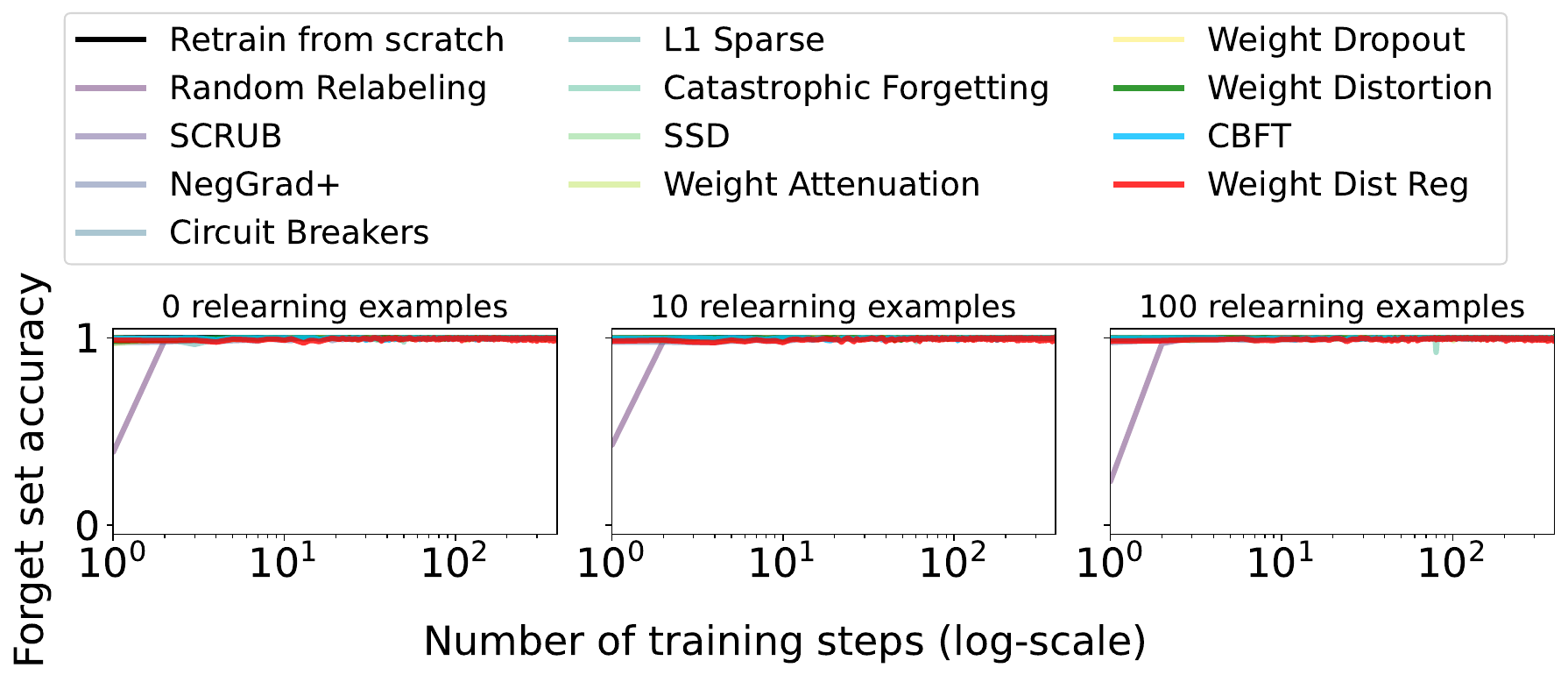}
    \includegraphics[width=\linewidth]{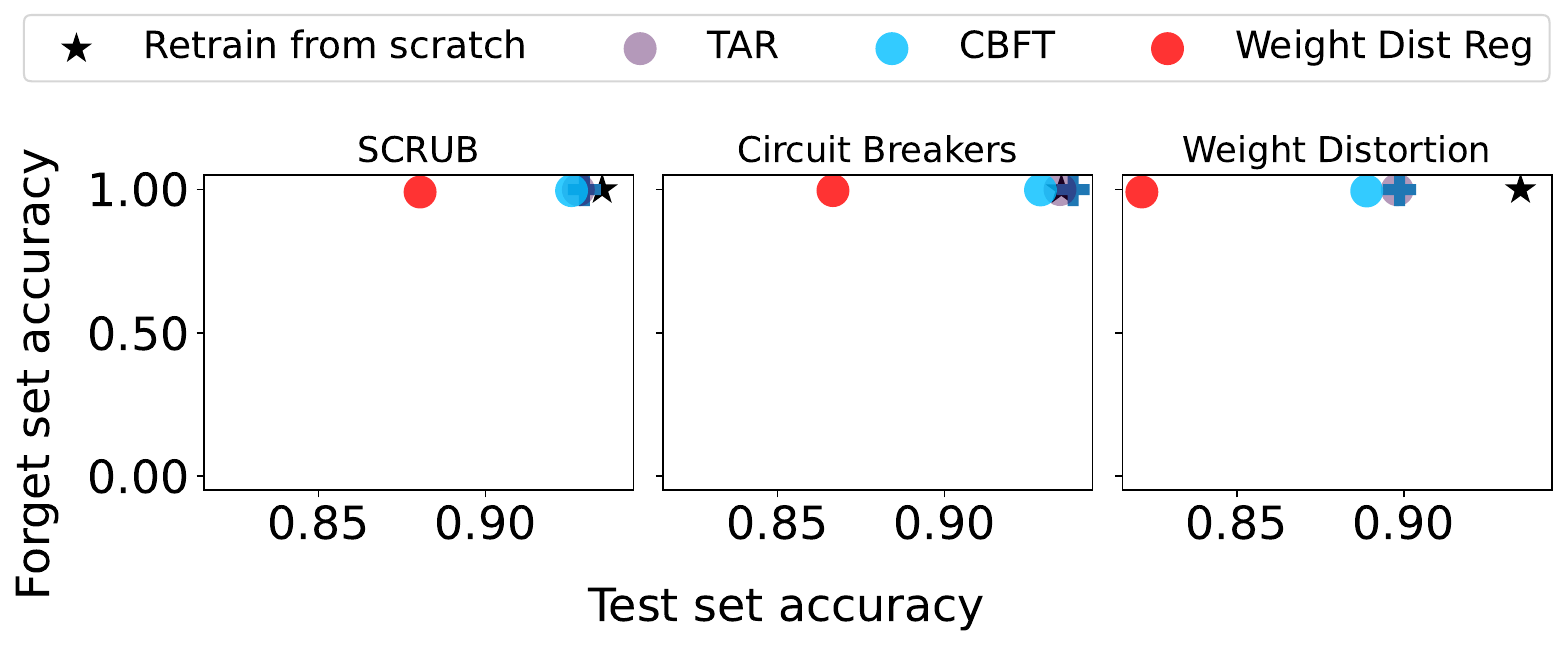}
    \vspace{-5mm}
    \caption{
    Comparison between test set accuracy and accuracy on the held-out part of the forget set $\mathcal{D}_{F_{ho}}$ on CIFAR-10 and ResNet-18, where the forget set is comprised of \textbf{typical examples} from the `airplane' class. The figure indicates that all methods achieve perfect recovery of unlearned knowledge, which is consistent with the retrain from scratch baseline, as the model can directly generalize to these examples without having them as part of the training set by definition.}
    \label{fig:relearning_forget_cifar10_resnet18_low_mem}
\end{figure}

As highlighted in \cref{sec:main_results}, typical examples can be particularly easy to unlearn as simply ignoring these examples during training (i.e., a no-op) yields similar predictions as the retrained from scratch model~\citep{jia2023l1sparse}. This is because they are, by definition, easy to predict regardless of whether they are part of the training set or not ~\citep{feldman2020neural,jiang2020characterizing,baldock2021deep,siddiqui2022metadata}. Generally, because typical examples are more likely to be predicted correctly, they are easier to relearn as well.
For the sake of completeness, we include the results for typical examples in \cref{fig:relearning_forget_cifar10_resnet18_low_mem}.
As evident from the figure, even the retrained from scratch model already achieves perfect accuracy on the forget set, highlighting that unlearning, and in turn, relearning are both trivial for typical examples.
Therefore, we primarily focused on atypical examples in \cref{sec:main_results}.
We see that methods such as Random Relabeling, while completing forgetting the forget set, deviates in a very significant way from the retrain from scratch baseline, which is what we compare against.

\section{Tamper-Resistance with Random Set of Examples for the Forget Set}
\label{app:random_ex}

\begin{figure}[t]
    \centering
    \includegraphics[width=\linewidth]{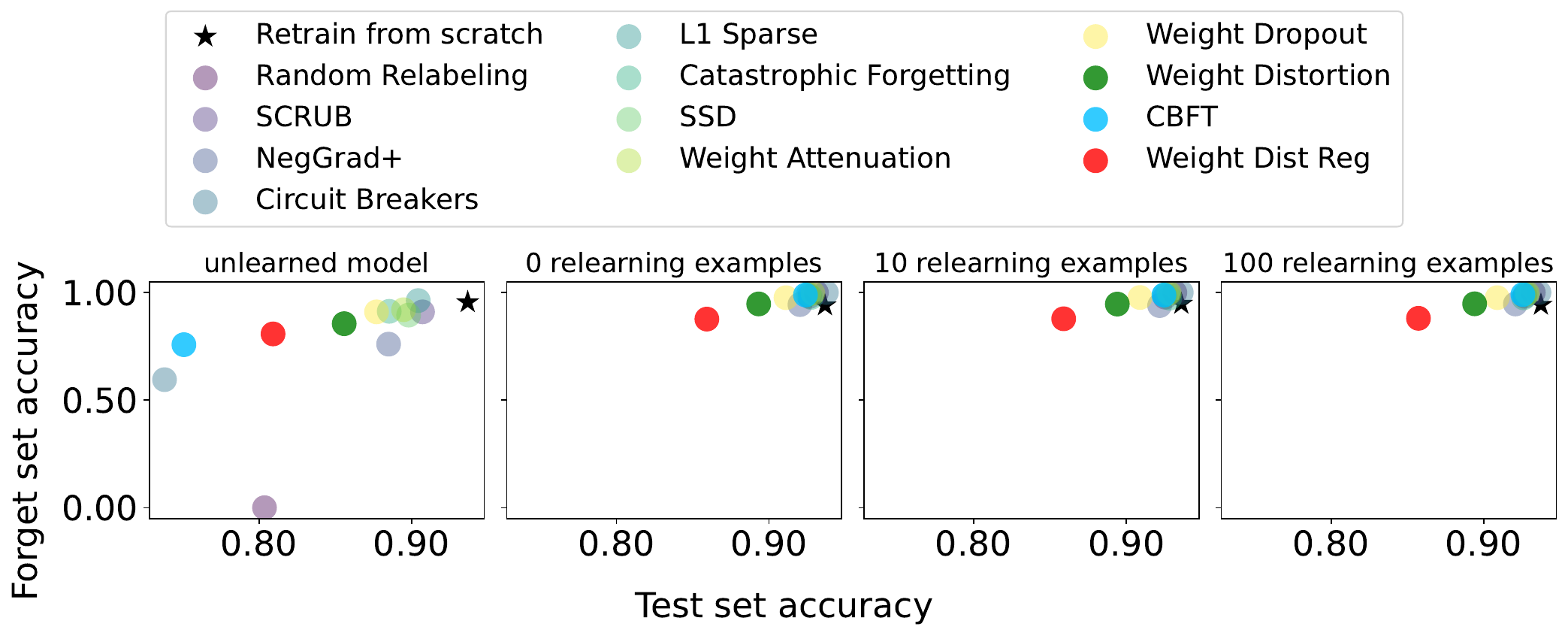}
    \includegraphics[width=\linewidth]{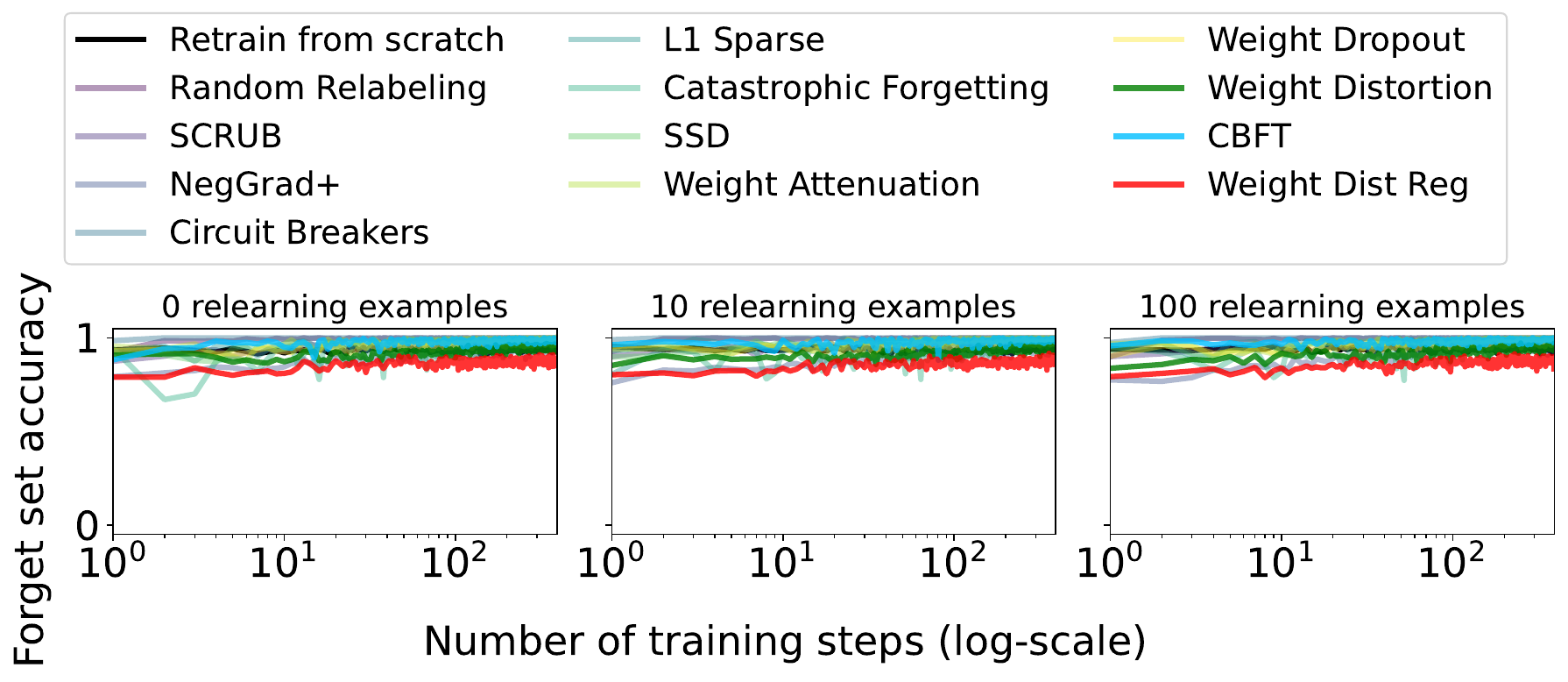}
    \includegraphics[width=\linewidth]{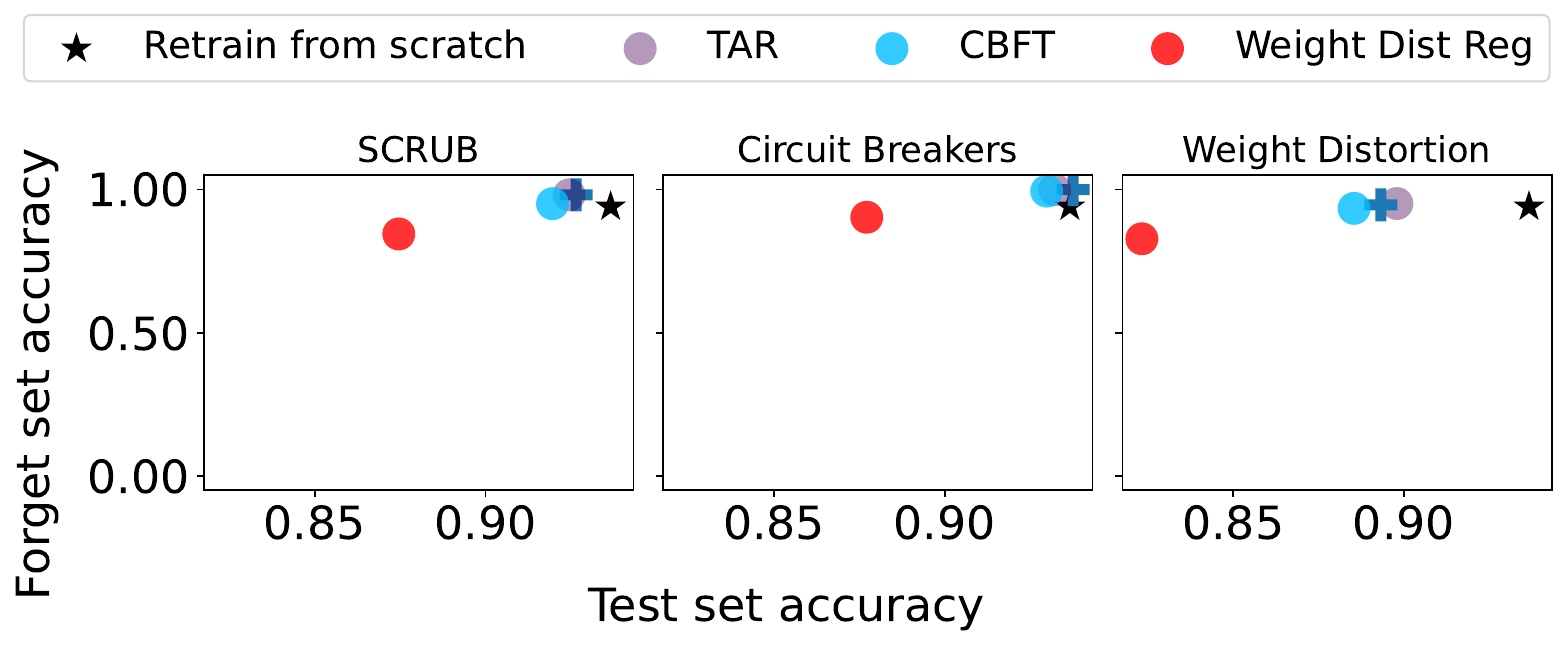}
    \vspace{-5mm}
    \caption{
    Comparison between test set accuracy and accuracy on the held-out part of the forget set $\mathcal{D}_{F_{ho}}$ on CIFAR-10 and ResNet-18, where the forget set is comprised of a \textbf{random subset} of the examples from the `airplane' class. The figure indicates that retrain from scratch baseline can already achieve near-perfect accuracy, as the examples in a class are predominantly comprised of typical examples.}
    \label{fig:relearning_forget_cifar10_resnet18_random}
\end{figure}

We further evaluate performance in the case of random selection of examples, rather than just typical examples as evaluated in \cref{app:typical_ex}.
We visualize these results in \cref{fig:relearning_forget_cifar10_resnet18_random}.
Since randomly selected examples are predominantly typical~\citep{feldman2020neural,feldman2020does,baldock2021deep,siddiqui2022metadata}, retrain from scratch already achieves near-perfect accuracy.
Furthermore, we observe only minor differences between methods.
Consequently, we focused on atypical examples for the forget set in the main paper (\cref{sec:main_results}) as these examples are hard to predict without being part of the training set, marking clearly the impact of relearning attacks which goes beyond model generalization.

\section{Tamper-Resistance on a More Complex Dataset (CIFAR-100)}
\label{app:cifar100}

\begin{figure}[t]
    \centering
    \includegraphics[width=\linewidth]{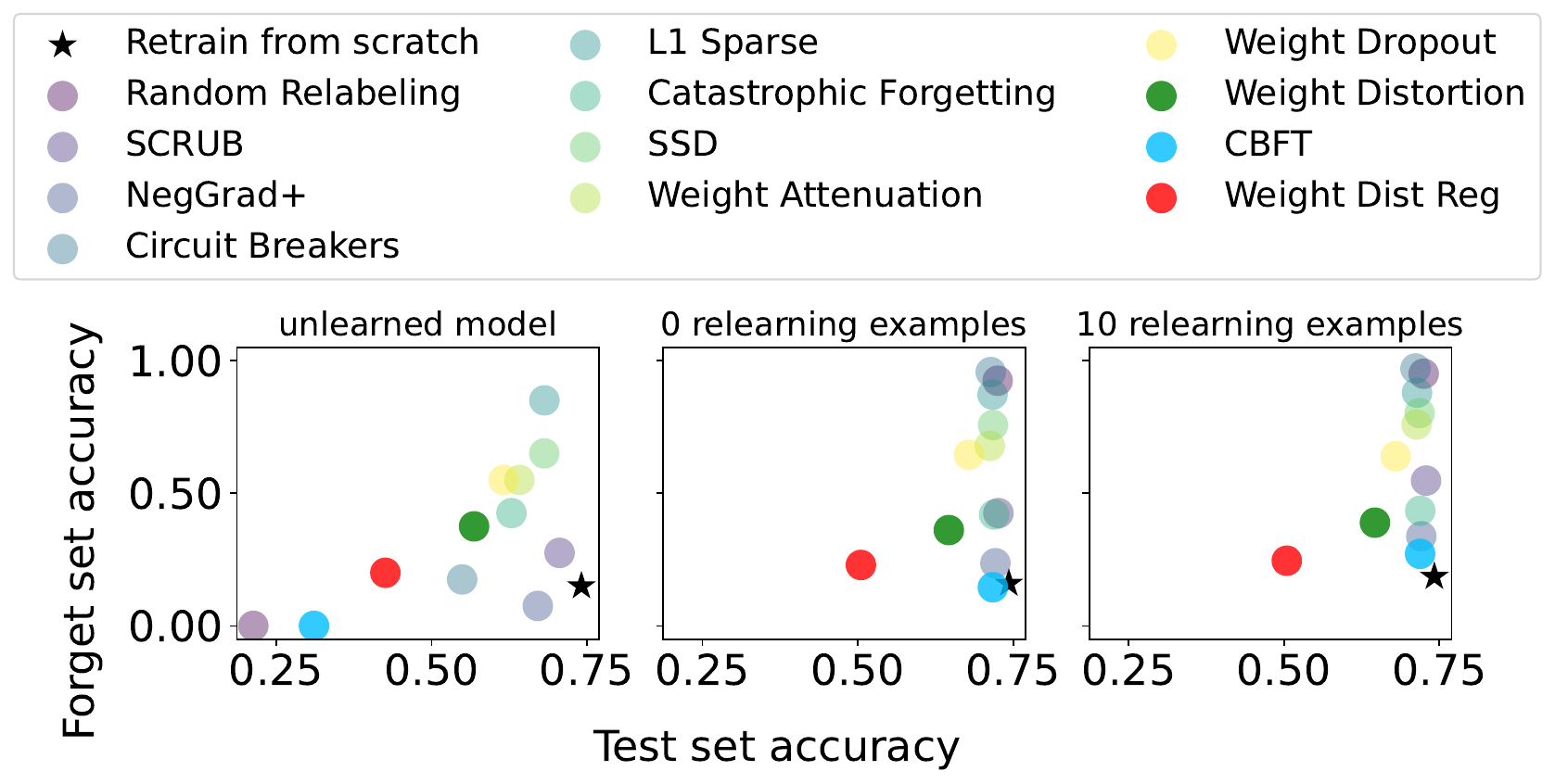}
    \includegraphics[width=\linewidth]{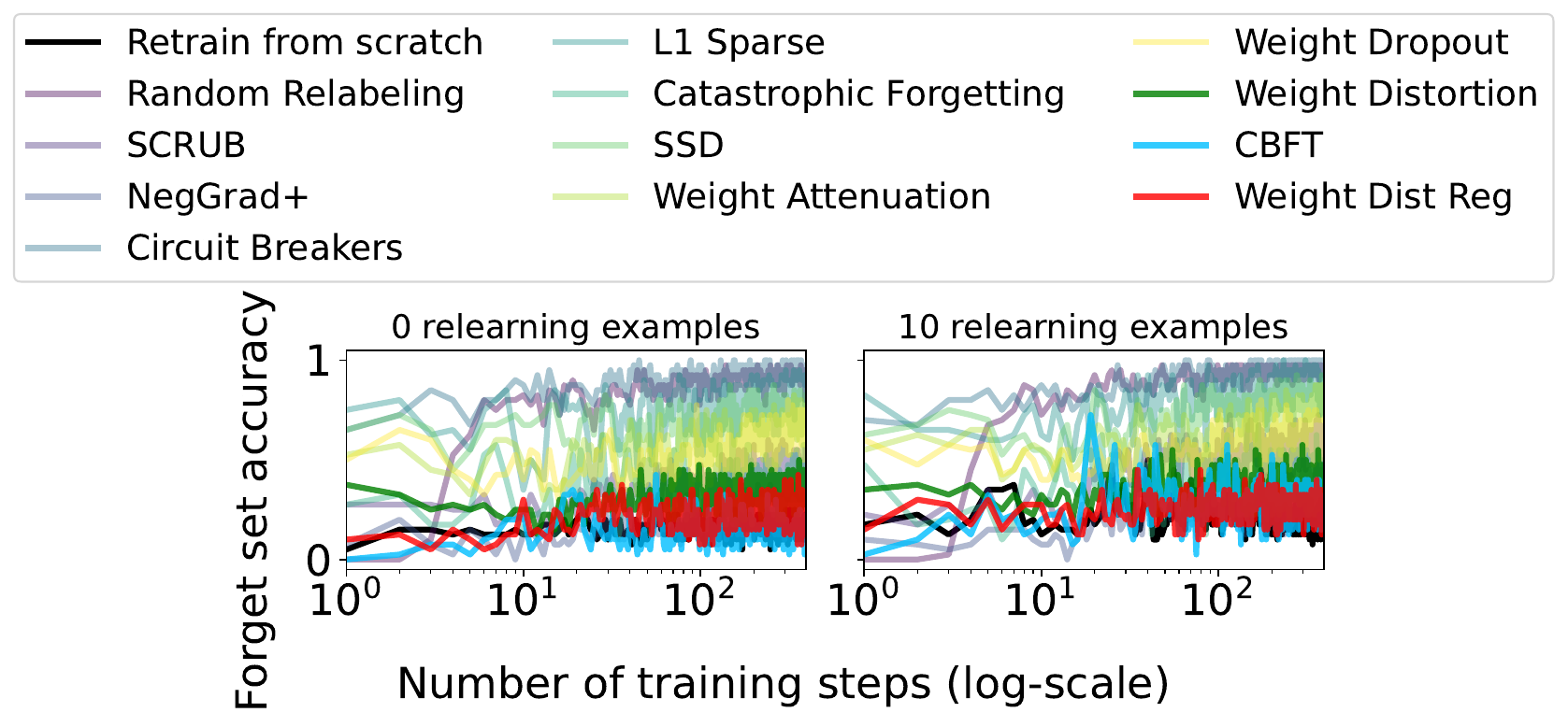}
    \includegraphics[width=0.9\linewidth]{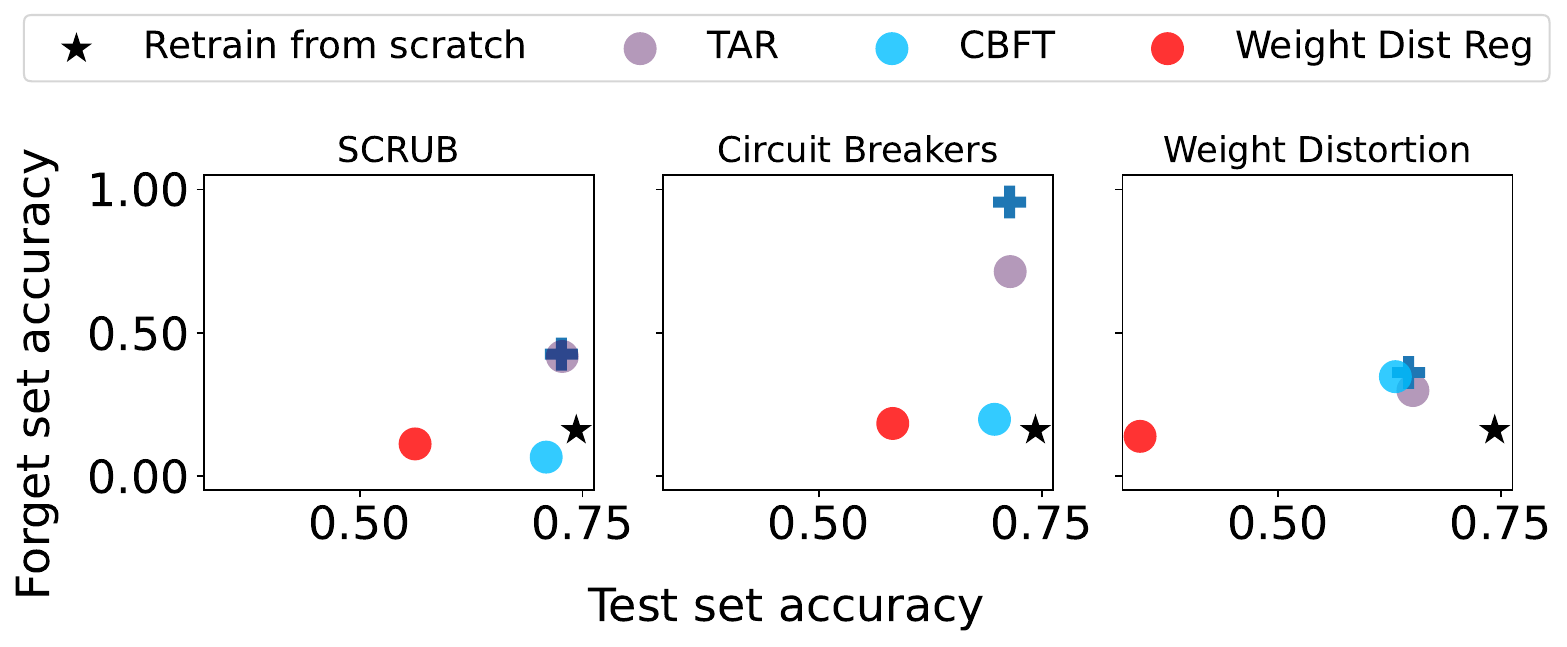}
    \caption{
    Comparison between test set accuracy and accuracy on the held-out part of the forget set $\mathcal{D}_{F_{ho}}$ on \textbf{CIFAR-100} and ResNet-18, where the forget set is comprised of atypical examples from the `apple' class.
    We see significant variation among different methods on the more complex CIFAR-100 dataset, which is comparable to the results we observed with class-agnostic unlearning on CIFAR-10 in \cref{fig:relearning_forget_cifar10_resnet18_high_mem} (bottom).
    }
    \label{fig:relearning_forget_cifar100_resnet18_high_mem}
\end{figure}

We predominantly focus on CIFAR-10 dataset in 
the main paper (\cref{sec:main_results}).
We evaluate the susceptibility on a more complex CIFAR-100 dataset in \cref{fig:relearning_forget_cifar100_resnet18_high_mem} in order to understand the impact of the dataset.
We only look at sub-class unlearning in this case, focusing on the `apple' class for unlearning.
Due to the higher complexity of the dataset, we observe lower test accuracies as evident from the scale of the x-axis.
Retrain from scratch baseline achieves a low forget set accuracy.
Furthermore, we see a wider spread of different methods on this more complex dataset, with the trend looking similar to the class-agnostic results presented in \cref{fig:relearning_forget_cifar10_resnet18_high_mem} (as it provided a more complex forget set comprising of the most atypical examples from the dataset).
It is interesting to note that CBFT~\citep{lubana2023mechanistic}, which was not tamper-resistant on CIFAR-10, demonstrated significant resistance on CIFAR-100.

\section{Tamper-Resistance on a Higher-Resolution Dataset (Imagenette)}
\label{app:imagenette}

\begin{table}
\centering
\begin{tabular}{l l r r}
\toprule
Model & Label & Test Acc (\%) & Forget Set Acc (\%) \\
\midrule
Unlearned & Retrain from Scratch & 87.34\% & 67.53\% \\
Unlearned & SCRUB & 85.81\% & 85.71\% \\
Unlearned & Weight Distortion & 83.16\% & 74.03\% \\
Unlearned & Weight Dist Reg & 83.72\% & 84.42\% \\
\addlinespace
0 Relearn & Retrain from Scratch & 86.40\% & 63.53\% \\
0 Relearn & SCRUB & 86.47\% & 93.40\% \\
0 Relearn & Weight Distortion & 84.71\% & 78.86\% \\
0 Relearn & Weight Dist Reg & 84.31\% & 73.87\% \\
\addlinespace
10 Relearn & Retrain from Scratch & 86.73\% & 64.16\% \\
10 Relearn & SCRUB & 86.65\% & 94.21\% \\
10 Relearn & Weight Distortion & 84.62\% & 81.09\% \\
10 Relearn & Weight Dist Reg & 84.83\% & 75.77\% \\
\bottomrule
\end{tabular}
\vspace{3mm}
\caption{Test and forget set accuracy across unlearning and relearning conditions when evaluating on the higher-resolution Imagenette dataset~\citep{imagenette} with atypical examples for the forget set. These results are primarily consistent with our findings on both CIFAR-10 as well as CIFAR-100 datasets, where models learned with our proposed unlearning algorithm are much more robust against relearning attacks.}
\label{tab:imagenette_results}
\end{table}

All our prior evaluations focused on small resolution images, i.e., $32 \times 32$ for both CIFAR-10 and CIFAR-100.
In order to understand the generalizability of our approaches to higher resolution images, we evaluate on the higher-resolution ImageNette dataset~\citep{imagenette}, which is a subset of the ImageNet dataset~\citep{ILSVRC15}.
The results are presented in \cref{tab:imagenette_results}.
Similarly to our findings on both CIFAR-10 (\cref{sec:main_results}) and CIFAR-100 (\cref{app:cifar100}), we see that the accuracy of \textit{retrain from scratch} on the forget set after relearning stays close to 50\%, while the accuracy of unlearning algorithms is significantly higher, except for our proposed \textit{weight dist reg} algorithm among the tested methods.

\section{Tamper-Resistance of a Larger Model}
\label{app:resnet34}

\begin{figure}[t]
    \centering
    \includegraphics[width=\linewidth]{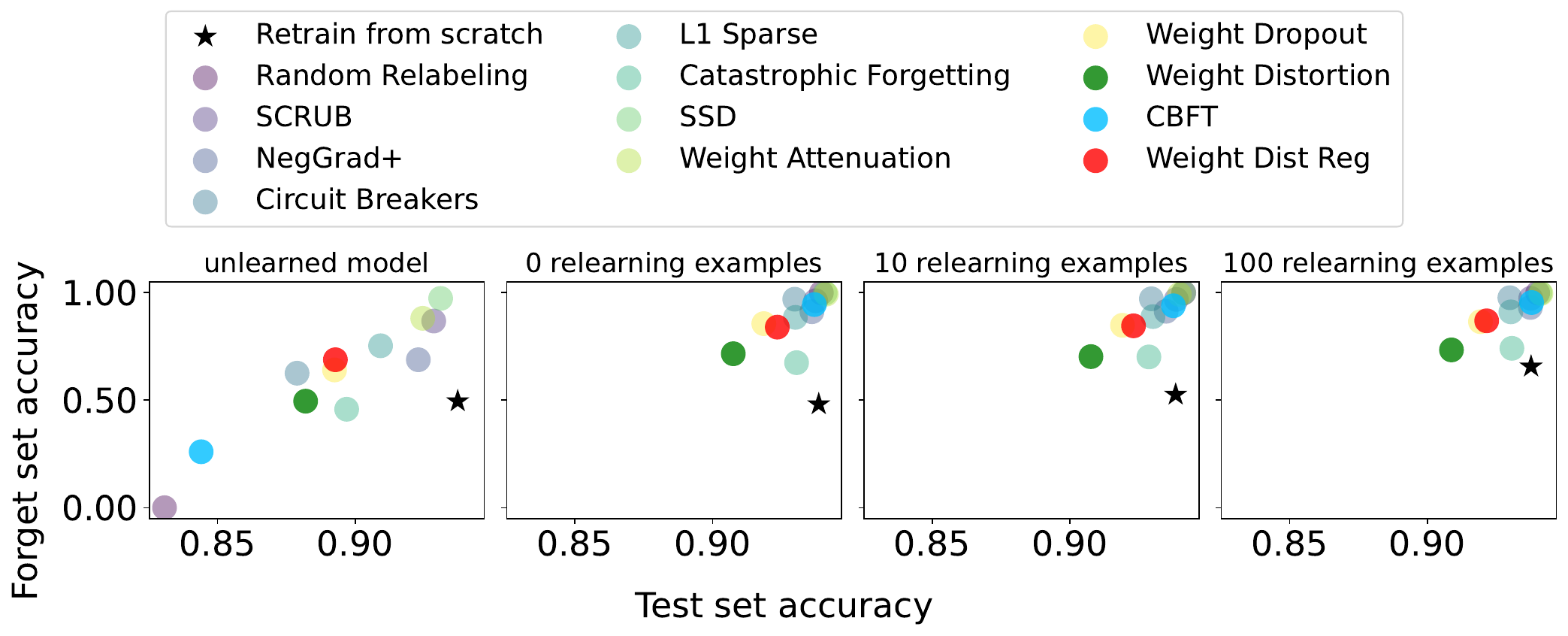}
    \includegraphics[width=\linewidth]{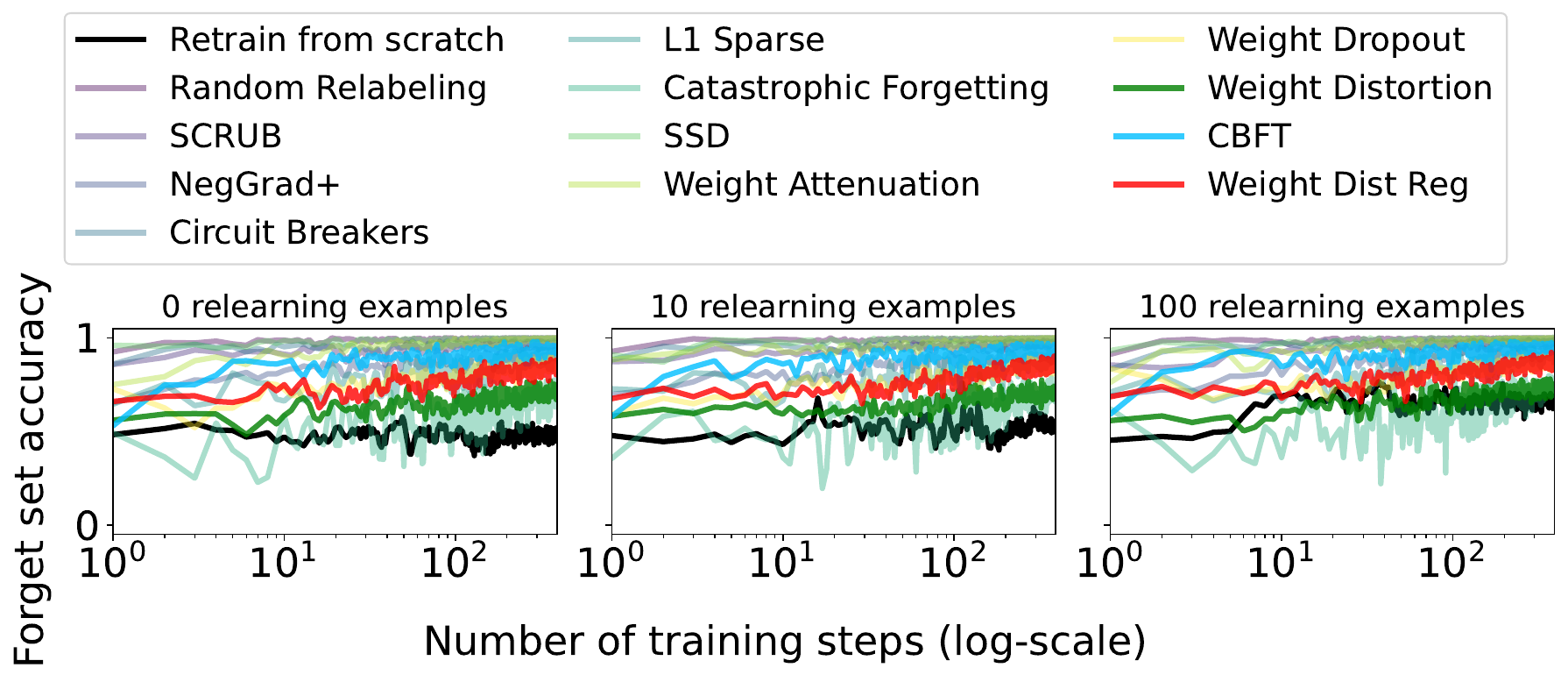}
    \includegraphics[width=\linewidth]{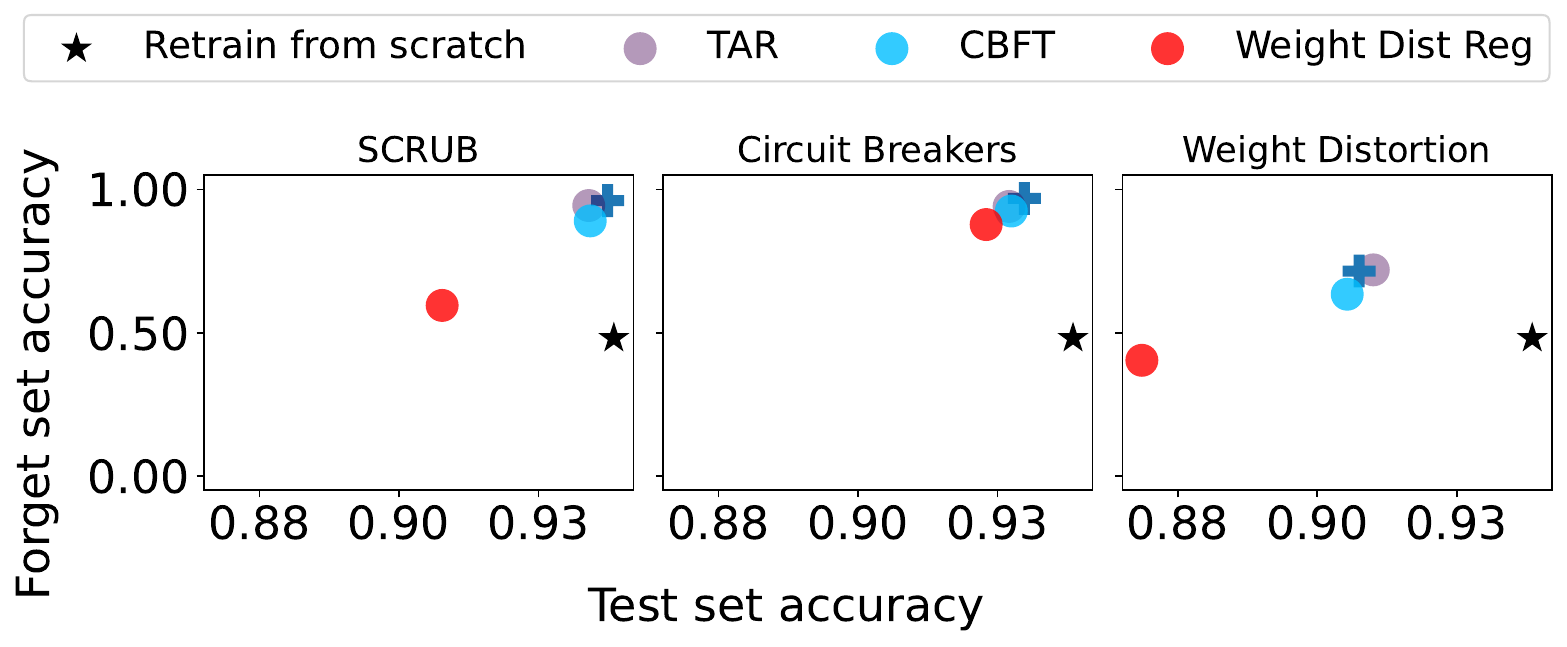}
    \vspace{-5mm}
    \caption{
    Comparison between test set accuracy and accuracy on the held-out part of the forget set $\mathcal{D}_{F_{ho}}$ on CIFAR-10 and \textbf{ResNet-34}, where the forget set is comprised of atypical examples from the `airplane` class. We still see the distinction between robust and non-robust methods. However, the relative ranking between different methods changes as we use the same set of hyperparameters, which should be tuned considering a larger number of parameters in the model. We find that using a large model, i.e., ResNet-34 instead of ResNet-18 induces a slight shift in relative method ranking, particularly for methods that require careful tuning of hyperparameters.}
    \label{fig:relearning_forget_cifar10_resnet34_high_mem}
\end{figure}

All our prior results focused on the smaller model, i.e., ResNet-18~\citep{he2016deep}.
In order to understand the impact of model size, we evaluate the susceptibility to relearning attacks on ResNet-34 (with almost double the number of parameters compared to ResNet-18, going from $\sim 11M$ to $\sim 21M$).
The results are visualized in \cref{fig:relearning_forget_cifar10_resnet34_high_mem}.
Note that we directly transfer the hyperparameter settings from our ResNet-18 experiments.
Hence, we found that Weight Dist Reg, which was the most competitive method in terms of robustness, became susceptible to relearning.
It is worth noting that the test set accuracy as well as the forget set accuracy for Weight Dist Reg in this case deviates significantly from all prior results, where it demonstrated consistently lower accuracies, highlighting a potential deficiency of the selected hyperparameters.
Catastrophic Forgetting and Weight Dist Reg exhibit high robustness against relearning, which are instantiations of our framework.

\section{Evolution of Forget Set Accuracy During Relearning}

\begin{figure}[t]
    \centering
    \includegraphics[width=\linewidth]{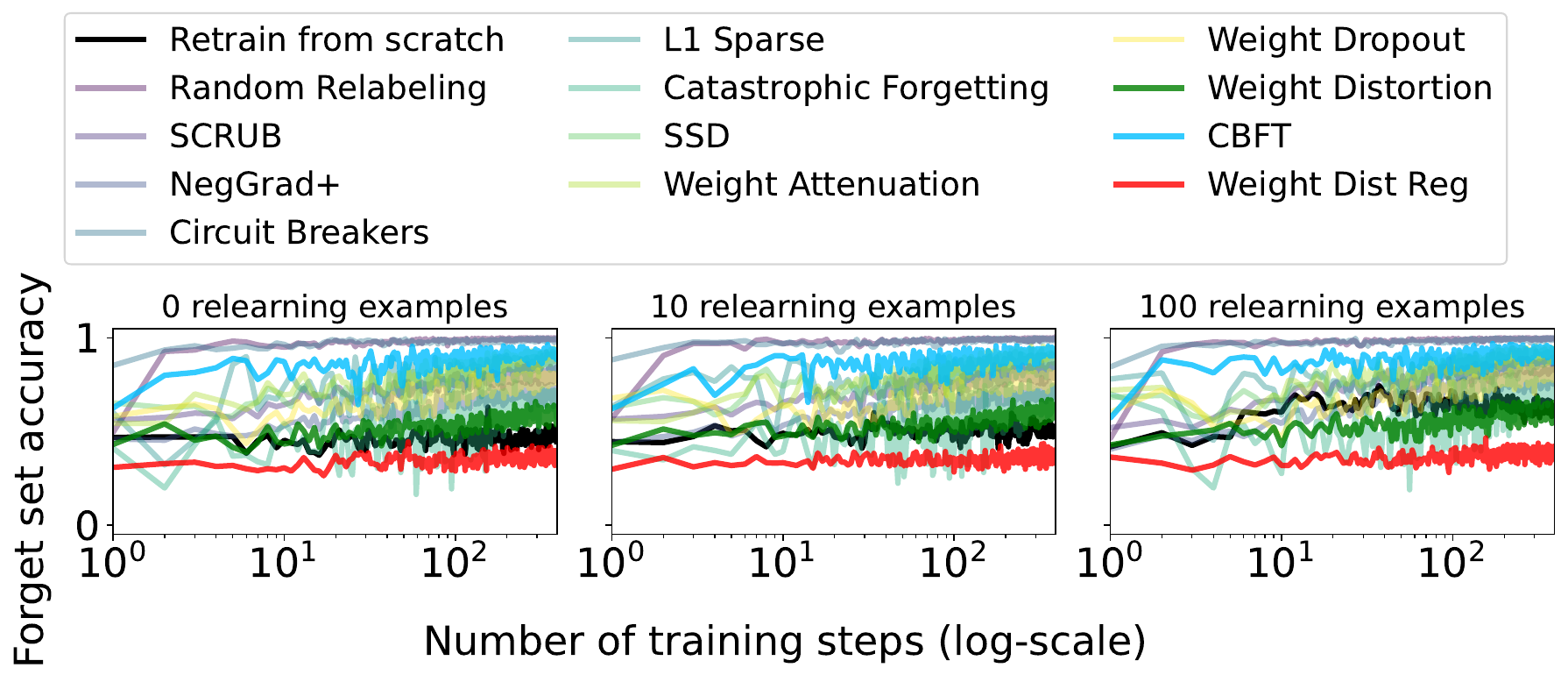}
    \vspace{-5mm}
    \caption{Comparison between relearning time and accuracy on the held-out part of the forget set $\mathcal{D}_{F_{ho}}$ on CIFAR-10 and ResNet-18, where the forget set is comprised of atypical examples from the `airplane' class. The figure indicates that many methods achieve near-perfect recovery of unlearned knowledge with only a small amount of model fine-tuning, without even assuming access to the unlearned examples. This figure is an extension of the results presented in \cref{fig:relearning_forget_cifar10_resnet18_high_mem} (top).}
    \label{fig:relearning_forget_cifar10_resnet18_high_mem_evolution}
\end{figure}

\begin{figure}[t]
    \centering
    \includegraphics[width=\linewidth]{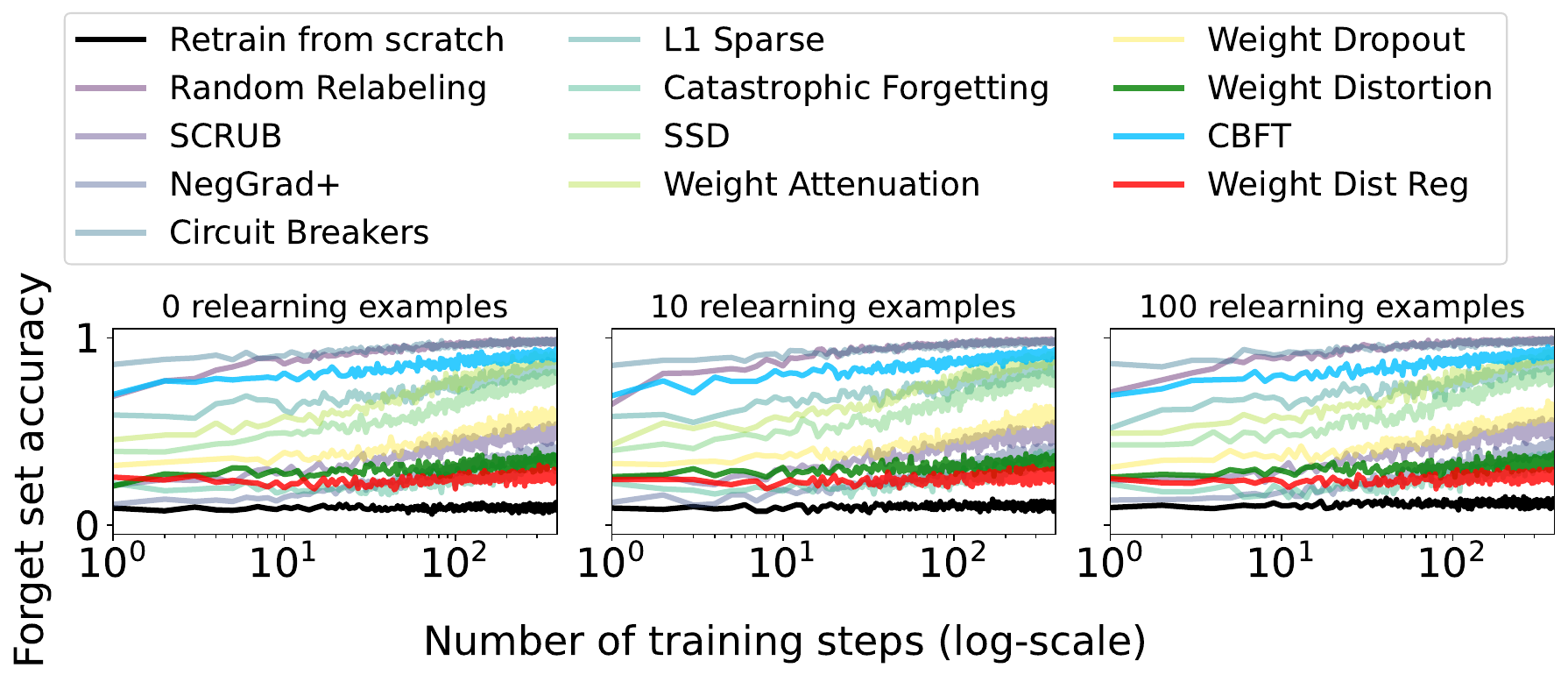}
    \includegraphics[width=\linewidth]{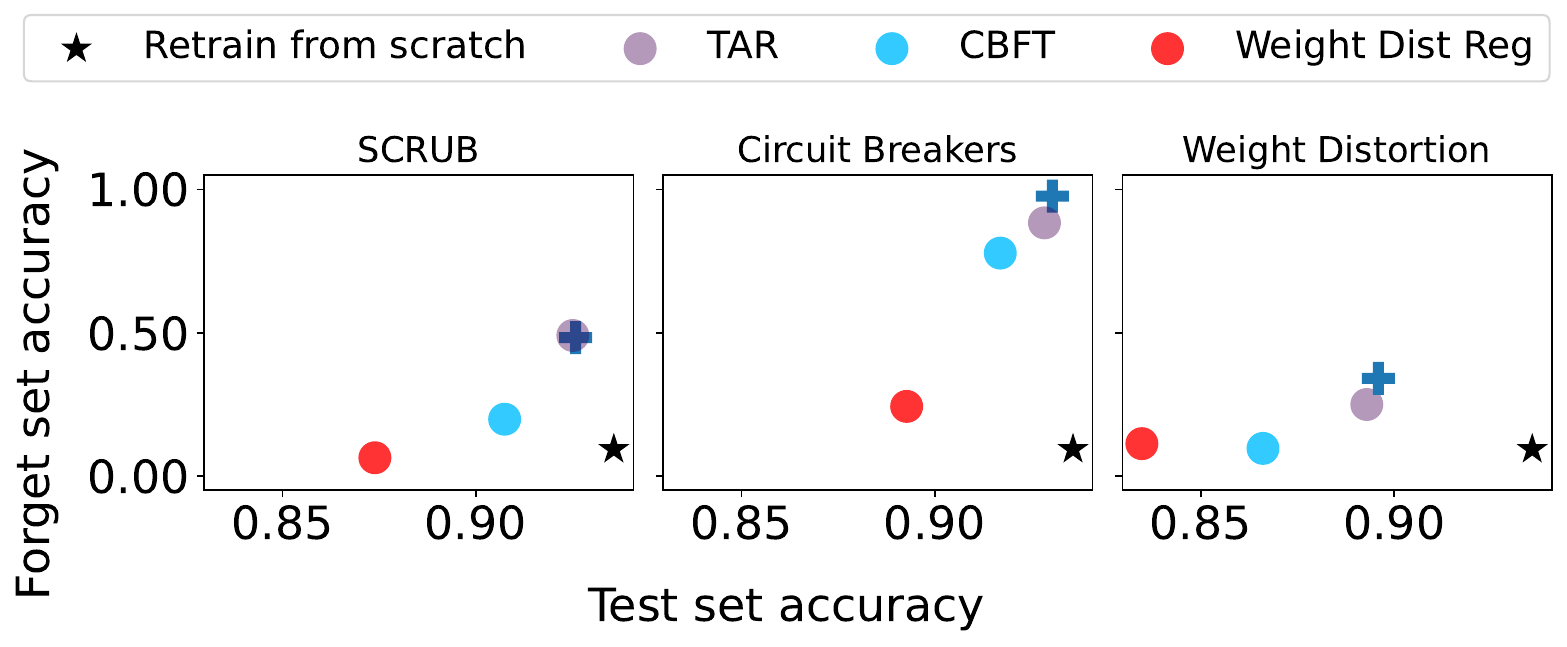}
    \vspace{-5mm}
    \caption{
    Comparison between test set accuracy and accuracy on the held-out part of the forget set $\mathcal{D}_{F_{ho}}$ on CIFAR-10 and ResNet-18, where the forget set is comprised of atypical examples from all classes. This figure is an extension of the results presented in \cref{fig:relearning_forget_cifar10_resnet18_high_mem} (bottom).
    We find that atypical examples, when selected from all the classes, are significantly harder in comparison to examples selected from a particular class, and hence, achieve a better separation between different methods in comparison to sub-class unlearning in \cref{fig:relearning_forget_cifar10_resnet18_high_mem} (top).
    }
    \label{fig:relearning_forget_cifar10_resnet18_high_mem_all_cls_remaining}
\end{figure}

In order to complement the results in \cref{fig:relearning_forget_cifar10_resnet18_high_mem} (top) and \cref{fig:relearning_forget_cifar10_resnet18_high_mem_new_init_sg}, we visualize the forget set accuracy evolution in \cref{fig:relearning_forget_cifar10_resnet18_high_mem_evolution}.
We see that methods that are most stable are nearly unaffected by increasing training time, highlighting that more training time might not be sufficient to increase their susceptibility further.

Similarly, \cref{fig:relearning_forget_cifar10_resnet18_high_mem_all_cls_remaining} complements the results on class-agnostic unlearning in \cref{fig:relearning_forget_cifar10_resnet18_high_mem} (bottom), where we visualize both the line plots as well as the results from two-phase training strategies.
We visually observe best separation between different methods on the line plot in this case of class-agnostic unlearning.

\section{Relearning with Examples from the Same Distribution, which are Distinct from the Retain Set}
\label{app:relearn_ex_type}

\begin{figure}[t]
    \centering
    \includegraphics[width=0.9\linewidth]{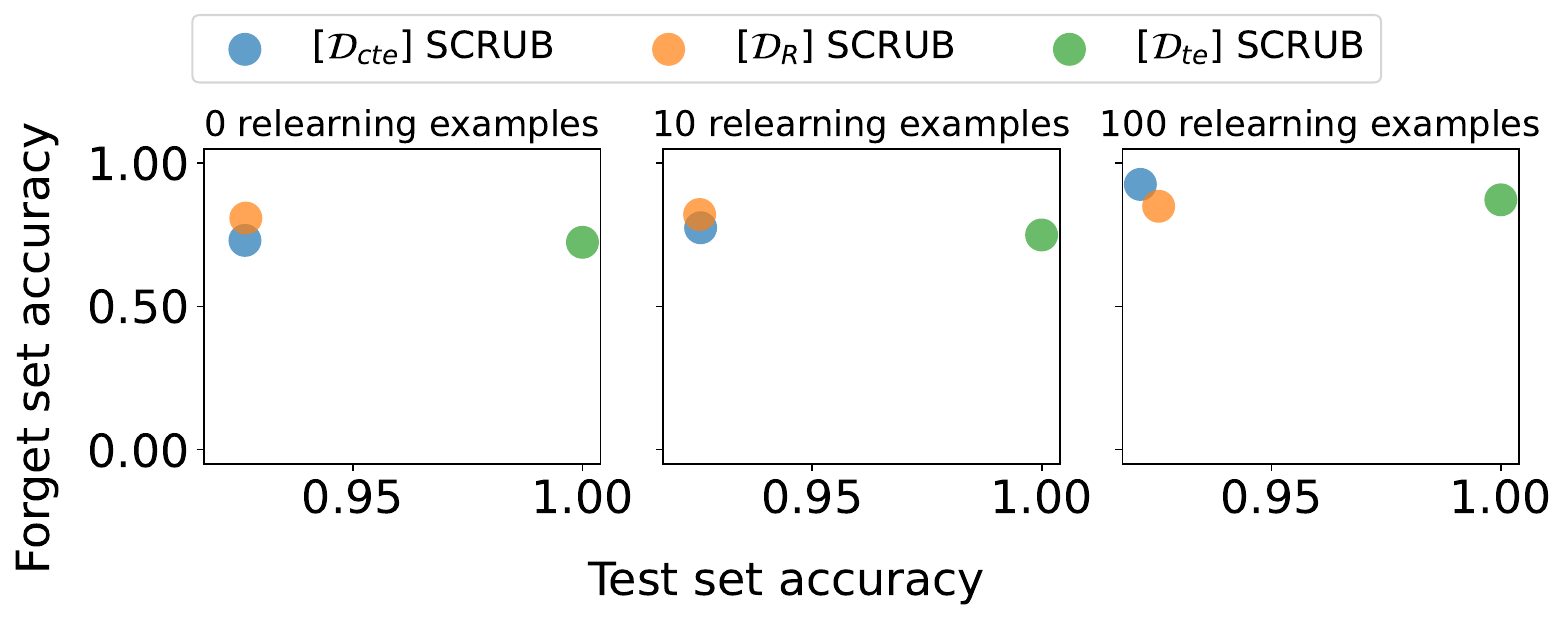}
    \includegraphics[width=\linewidth]{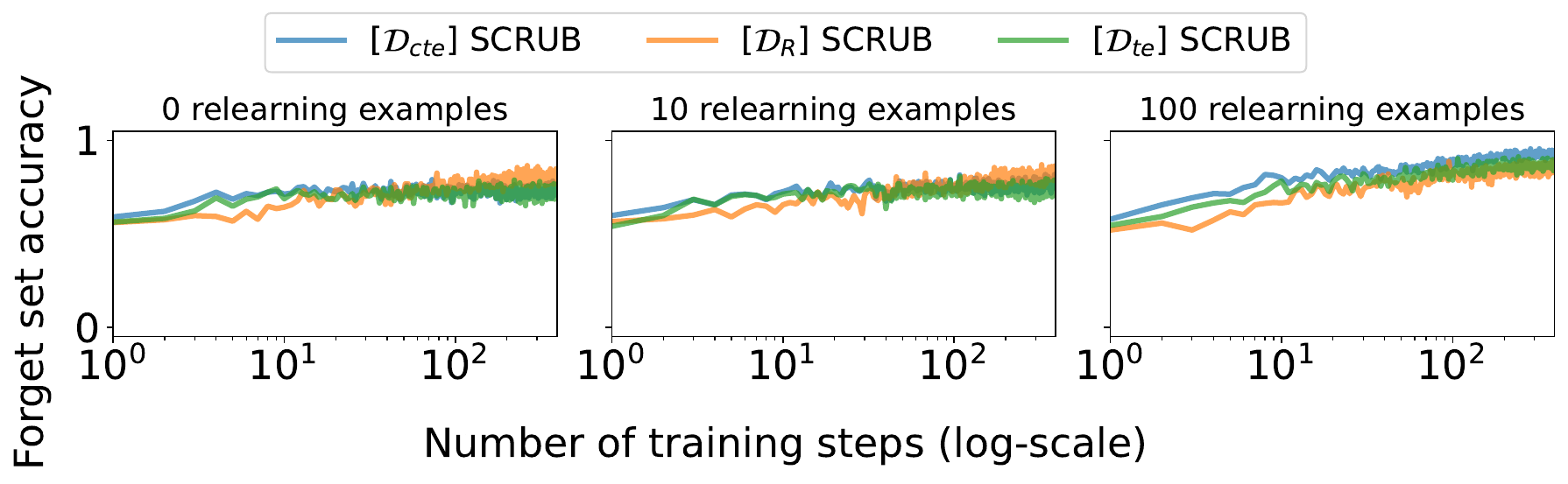}
    \vspace{-3mm}
    \caption{Comparison between training time and accuracy on the held-out part of the forget set $\mathcal{D}_{F_{ho}}$ on CIFAR-10 and ResNet-18, where the forget set is comprised of atypical examples, and using SCRUB as the unlearning method.
    We evaluate the use of different sets to remind the model, including the retain set $\mathcal{D}_{R}$, test set $\mathcal{D}_{te}$, corrupted test set $\mathcal{D}_{cte}$ taken from CIFAR-10-C~\citep{hendrycks2019benchmarking} (using JPEG compression with a severity level of 5), along with the selected number of relearning examples.
    The figure indicates that there can be significant differences in relearned model accuracy based on the selected examples used for reminding the model.}
    \label{fig:relearning_different_sets}
\end{figure}

While we demonstrated that unlearned knowledge can be recovered through repeated fine-tuning on the retain set, it remains unclear whether we can also recover performance on the forget set by fine-tuning on new examples from the same data distribution that are unseen by the model.
The question that we ask is: \textit{are the same examples that the model was trained on particularly important for relearning, or other examples from the same data distribution equally effective?}

In order to simulate this scenario, we use examples from the test set (which are unseen by the model) for relearning instead of the retain set.
We further evaluate using a corrupted version of the test set from CIFAR-10-C~\citep{hendrycks2019benchmarking}.
In particular, we use the JPEG corrupted examples with the highest severity level of 5.
Note that we still use examples from the relearning set when considering scenarios where the number of relearning examples is > 0.

The results are presented in \cref{fig:relearning_different_sets}. We observe that using the retain set $\mathcal{D}_{R}$ achieves higher accuracy on the forget set compared to other choices. However, these differences diminish as the number of relearning examples increases, since these examples directly represent the unlearned knowledge and may become the dominant factor in recovery. When the test set is used for relearning (instead of the retain set), we unsurprisingly observe perfect accuracy on the test set, as it serves as the training set in this scenario. These results indicate that relearning is more effective when using examples that were seen during training, rather than a held-out set of examples.

These findings are also related to the other recent findings on anticipatory knowledge recovery when exposing the model to a repeated sequence of documents. In this scenario, as the model processes documents in a fixed order, it unexpectedly begins to recover an increasing amount of information about a previously seen example even before encountering that example again~\citep{yang2024reawakening}.

\section{Membership-Inference Attacks (MIA) on Unlearned Models}
\label{app:mia}

\begin{table}
\centering
\begin{tabular}{l r}
\toprule
Unlearning Method & MIA Acc (\%) \\
\midrule
Retrain from Scratch & 52.2\% \\
L1 Sparse & 54.2\% \\
SCRUB & 51.6\% \\
Circuit Breakers & 55.0\% \\
Gradient Ascent & 54.2\% \\
Random Relabeling & 94.8\% \\
Catastrophic Forgetting & 54.2\% \\
SSD & 53.8\% \\
Weight Attenuation & 53.6\% \\
Weight Distortion & 53.2\% \\
Weight Dropout & 54.8\% \\
Weight Dist Reg & 48.4\% \\
\bottomrule
\end{tabular}
\vspace{3mm}

\caption{Membership-Inference Attack (MIA) accuracy, where 50\% indicates chance-level performance. Note that most of the methods are close to chance-level performance, except random relabeling. This indicates that MIA attacks are not sufficient to understand susceptibility to relearning attacks.}
\label{tab:mia_acc}
\end{table}

Membership-Inference Attacks (MIA) attempt to evaluate if it is possible to identify instances that were part of the model's training distribution from held-out instances~\citep{shokri2017membership}.
Being able to distinguish instances leaks private information regarding an instance's membership.
We follow the setup from \citep{kurmanji2024scrub} for membership-inference attacks with balanced loss-threshold-based classifier.
The results are presented in \cref{tab:mia_acc}.
We observe that, as is the case in our initial findings with the forget set accuracy metric, this particular type of MIA can give a false sense that unlearning successfully erased all traces of unlearned data, even in cases where we know that the unlearned model is prone to relearning attacks.
Note that we evaluated a particularly simple variant of MIA based on prior work~\citep{kurmanji2024scrub}. Therefore, it is possible to use more complex variants of MIA, which might be better able to predict this vulnerability against relearning.

\section{Understanding the Efficiency of Unlearning Algorithms}
\label{app:unlearning_efficiency}

\begin{table}
\centering
\begin{tabular}{l l r r r}
\toprule
Model & Method & Unlearning Epochs & Test Acc (\%) & Forget Set Acc (\%) \\
\midrule
Unlearned & Catastrophic Forgetting & 1 & 93.79\% & 100.00\% \\
Unlearned & Catastrophic Forgetting & 3 & 93.42\% & 100.00\% \\
Unlearned & Catastrophic Forgetting & 10 & 92.52\% & 94.50\% \\
Unlearned & Catastrophic Forgetting & 30 & 91.68\% & 77.75\% \\
Unlearned & Catastrophic Forgetting & 50 & 90.25\% & 60.00\% \\
Unlearned & Catastrophic Forgetting & 100 & 88.23\% & 41.75\% \\
Unlearned & Weight Distortion & 1 & 84.49\% & 43.50\% \\
Unlearned & Weight Distortion & 3 & 86.21\% & 49.00\% \\
Unlearned & Weight Distortion & 10 & 86.27\% & 52.75\% \\
Unlearned & Weight Distortion & 30 & 86.29\% & 53.25\% \\
Unlearned & Weight Distortion & 50 & 86.31\% & 50.50\% \\
Unlearned & Weight Distortion & 100 & 85.97\% & 45.75\% \\
\addlinespace
0 Relearn & Catastrophic Forgetting & 1 & 93.71\% & 99.99\% \\
0 Relearn & Catastrophic Forgetting & 3 & 93.65\% & 99.99\% \\
0 Relearn & Catastrophic Forgetting & 10 & 93.37\% & 99.28\% \\
0 Relearn & Catastrophic Forgetting & 30 & 93.22\% & 93.63\% \\
0 Relearn & Catastrophic Forgetting & 50 & 93.08\% & 82.16\% \\
0 Relearn & Catastrophic Forgetting & 100 & 92.80\% & 66.14\% \\
0 Relearn & Weight Distortion & 1 & 90.14\% & 61.07\% \\
0 Relearn & Weight Distortion & 3 & 90.14\% & 61.66\% \\
0 Relearn & Weight Distortion & 10 & 89.86\% & 61.38\% \\
0 Relearn & Weight Distortion & 30 & 89.62\% & 60.69\% \\
0 Relearn & Weight Distortion & 50 & 89.44\% & 59.64\% \\
0 Relearn & Weight Distortion & 100 & 89.64\% & 58.59\% \\
\bottomrule
\end{tabular}
\vspace{2mm}
\caption{Test and forget-set accuracy of the unlearned model after relearning attack on CIFAR-10 with atypical forget set examples. It is evident from the results that even a small number of unlearning epochs are sufficient to achieve robustness against relearning attack with \textit{weight distortion}. In contrast, other methods such as \textit{catastrophic forgetting} only work at the largest epoch setting. Note that all previous results were reported with 100 unlearning epochs, masking any impact of the efficiency of the unlearning algorithm.}
\label{tab:unlearning_efficiency}
\end{table}

In order to control for efficiency effects of the unlearning algorithm, all previous results were reported at a fixed unlearning budget of 100 epochs, which is comparable to the budget allocated for pretraining.
This raises concerns regarding the utility of approximate unlearning algorithms, as one should use the gold-standard unlearning algorithm, i.e., \textit{retrain from scratch} in cases where the computational budget is directly comparable to the pretraining budget.

In order to understand the significance of this selection, we vary the number of unlearning epochs and report test accuracy and forget set accuracy after relearning, while using the same set of hyperparameters.
The results are presented in \cref{tab:unlearning_efficiency}.
It is clear from the table that while both catastrophic forgetting and weight distortion are successful in resisting relearning at the maximum budget of 100 epochs, catastrophic forgetting is still susceptible to relearning with a smaller number of unlearning epochs. On the other hand, weight distortion is more resistant to relearning even after a single epoch of unlearning.

\section{Computational Resources}
\label{app:comp_resouces}

We used NVIDIA RTX 3090 for each of our experiments, with the GPU equipped with 24GB of high-bandwidth memory (HBM) -- we only use a tiny fraction of it as we train small ResNet models on CIFAR-10/100.

The pretraining takes about 3 hours.
Each unlearning method takes about 2.5 hours, except TAR which takes about 9 hours.
Each setting required training 21 unlearned models ($3 \times 3$ combinations for two-phase training as we evaluate three methods with three different initial safeguards).
This equates to about $18 \times 2.5 + 9 \times 3 = 72$ hours per setting.
We evaluated 6 main settings: ResNet-18 on CIFAR-10 (typical, atypical, and random subset for the forget set), ResNet-34 on CIFAR-10 (atypical), ResNet-18 on CIFAR-10 with class-agnostic unlearning (atypical), and ResNet-18 on CIFAR-100 (atypical).
Hence, this equates to about 450 hours.
Grid evaluation of each model further takes one hour, resulting in an additional 130 hours ($21 \times 6$ models).

Finally, taking into account all further evaluations as well as failed experiments, we expect to have invested about 1000 GPU hours on the project.

\section{Societal Impact}
\label{app:soc_impact}

We highlight limitations of existing unlearning techniques in the context of visual recognition, showing that such techniques are prone to relearning attacks, where unlearned knowledge can be easily recovered with access to only retained knowledge. Deployment of such methods therefore poses risks in the context of privacy or model safety.
We further propose a new class of methods that are more robust against such attacks.

Developing better tamper-resistance to relearning attacks for unlearned models attempts to reduce the potential harm from these existing systems.

\end{document}